\documentclass[times, 10pt]{elsarticle}

\usepackage{lineno}
\usepackage{amssymb,amsmath}

\usepackage{amssymb,amsmath}
\usepackage{soul}
\usepackage{graphics}
\usepackage{psfig}
\usepackage{epsfig}
\usepackage{graphicx}
\usepackage{algorithmic}
\usepackage[ruled]{algorithm2e}
\usepackage{morefloats}
\usepackage{color,soul}
\usepackage{url}
\usepackage{mathtools, cuted}
\usepackage[most]{tcolorbox}
\usepackage{epsfig}
\usepackage{adjustbox}
\usepackage{pifont}
\newcommand{\cmark}{\ding{51}}%
\newcommand{\xmark}{\ding{55}}%
\usepackage{xcolor}
\usepackage{multirow}
\usepackage{color, colortbl}
\usepackage{amssymb}
\usepackage{amsmath}  
\usepackage{subcaption}
\usepackage{sidecap}
\usepackage{graphicx}
\usepackage{booktabs} 
\usepackage{xspace}

\makeatletter
\DeclareRobustCommand\onedot{\futurelet\@let@token\@onedot}
\def\@onedot{\ifx\@let@token.\else.\null\fi\xspace}

\def\eg{\emph{e.g}\onedot} 
\def\ie{\emph{i.e}\onedot} 
 
\def\etc{\emph{etc}\onedot} 
 
\def\etal{\emph{et al}\onedot}
\makeatother
\DeclareMathOperator*{\argmaxA}{arg\,max}
\usepackage[breaklinks,colorlinks]{hyperref}

\definecolor{lightcyan}{rgb}{0.88, 1.0, 1.0}
\definecolor{Gray}{gray}{0.94}
\definecolor{ForestGreen}{RGB}{34,139,34}
\definecolor{LightCyan}{RGB}{224, 255, 255}
\definecolor{BrickRed}{rgb}{.72,0,0} 
\definecolor{commentcolor}{RGB}{110,154,155}

\newcommand{\method}{{CleanAdapt}}
\newcommand{\methodj}{{CleanAdapt + TS}}

\begin{document}

\begin{frontmatter}

% \title{Overcoming Label Noise for \\Source-free Unsupervised Video Domain Adaptation}
%\title{Source-free Video Domain Adaptation as \\Learning from Noisy Labels}
\title{Source-free Video Domain Adaptation by \\Learning from Noisy Labels}
\author[1]{Avijit Dasgupta \corref{mycorrespondingauthor}}
\cortext[mycorrespondingauthor]
{Corresponding author}
\ead{avijit.dasgupta@research.iiit.ac.in}
\author[1]{C. V. Jawahar}
\ead{jawahar@iiit.ac.in}
\author[2]{Karteek Alahari}
\ead{karteek.alahari@inria.fr}
\address[1]{CVIT, IIIT Hyderabad, India}
\address[2]{Univ. Grenoble Alpes, Inria, CNRS, Grenoble
INP, LJK, France}

\begin{abstract}
Despite the progress seen in classification methods, current approaches for handling videos with distribution shifts in source and target domains remain source-dependent as they require access to the source data during the adaptation stage. In this paper, we present a self-training based \textit{source-free} video domain adaptation approach to address this challenge by bridging the gap between the source and the target domains. We use the source pre-trained model to generate pseudo-labels for the target domain samples, which are inevitably noisy. Thus, we treat the problem of source-free video domain adaptation as learning from noisy labels and argue that the samples with correct pseudo-labels can help us in adaptation. To this end, we leverage the cross-entropy loss as an indicator of the correctness of the pseudo-labels and use the resulting small-loss samples from the target domain for fine-tuning the model. We further enhance the adaptation performance by implementing a teacher-student (TS) framework, in which the teacher, which is updated gradually, produces reliable pseudo-labels. Meanwhile, the student undergoes fine-tuning on the target domain videos using these generated pseudo-labels to improve its performance. Extensive experimental evaluations show that our methods, termed as \textit{\method{}, \methodj{}}, achieve state-of-the-art results, outperforming the existing approaches on various open datasets. Our source code is publicly available at \url{https://avijit9.github.io/CleanAdapt}.
\end{abstract}


\end{frontmatter}

%-----------Intro-----------
\section{Introduction}
\label{sec:intro}

 % Deep neural networks with the rise of large-scale action recognition datasets have proven to be beneficial for the progress in video understanding~\cite{simonyan2014two,wang2016temporal,carreira2017quo,neimark2021video}. 
 
The availability of large-scale action recognition datasets, coupled with the rise of deep neural networks, have significantly advanced the field of video understanding~\cite{carreira2017quo}.
Similar to other machine learning models, these action recognition models often encounter new domains with \textit{distribution-shift} when deployed in real-world scenarios where the data distribution of training (source domain) and test (target domain) data is different, resulting in degraded performance. A trivial solution to alleviate this problem is fine-tuning the models with \textit{labeled} target domain data, which is not always feasible due to expensive target domain annotations. Unsupervised domain adaptation (UDA) tackles this problem by transferring knowledge from the labeled source domain data to the \textit{unlabeled} target domain, thus eliminating the need for comprehensive annotations for the target domain~\cite{ganin2015unsupervised}. Source-free UDA takes this approach one step further, where we assume that the source domain data is unavailable during the adaptation stage. This setting is more realistic than the source-dependent one primarily due to (a) privacy
concerns, it is not always possible to transfer data between the vendor (source) and the client (target), (b) storage constraints to transfer the source data to the client side (e.g., Sports-1M is about 1 TB), and (c) source-free
 models reduce computation time and thus cost by not
 using the source domain data during the adaptation stage.

% Action recognition models~\cite{simonyan2014two,wang2016temporal,carreira2017quo,neimark2021video} often encounter new domains with \textit{distribution-shift}~\cite{torralba2011unbiased} when deployed in the real world. Such shifts can occur in videos for several reasons: relative differences in the speed and duration of the action, camera movement, viewpoints, etc. Thus, the resulting difference in data distributions of the training (source domain) and the test (target domain) data produces a degraded performance. Furthermore, the source domain data usually comes with fully labeled videos, whereas the target domain data is typically unlabeled to reduce the annotation cost. Unsupervised domain adaptation (UDA) aims at adapting the model to the \textit{label-scarce} target domain by transferring the knowledge learned from the \textit{label-rich} source domain data~\cite{ganin2015unsupervised, long2015learning, saito2018maximum, wang2020classes, yang2021st3d}. Source-free UDA~\cite{liang2020we, yang2021exploiting, kundu2020universal} takes this approach one step further by assuming the unavailability of the source domain data for adaptation. This is a more practical setup than traditional source-dependent UDA mainly due to  privacy issues, computation cost, and storage complexity~\cite{huang2021model, liang2020we, yang2021exploiting}.

\begin{figure}[t]
    \centering
    \includegraphics[width=\columnwidth]{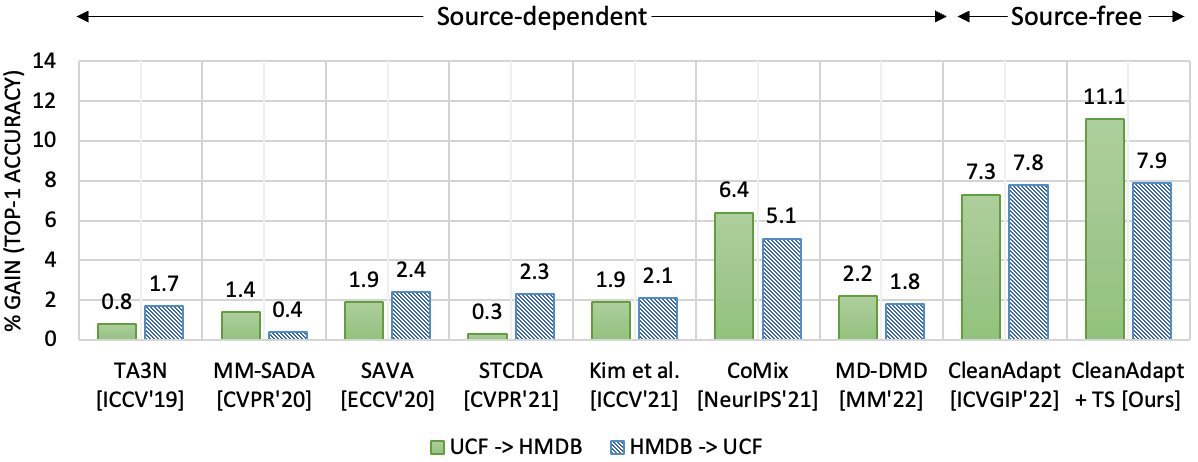}
    \caption{Existing approaches have a \textit{source-dependent} adaptation stage achieving marginal performance gain over the source-pretrained models. On the other hand, our proposed methods \method{} and \methodj{} achieve significant performance improvements over the source-only model while being \textit{source-free} (\ie, the adaptation stage does not require videos from the source domain). (Best viewed in color.)}
      \label{fig:teaser}
      
\end{figure}

There has been a recent surge of interest in source-dependent unsupervised domain adaptation for videos~\cite{kim2021learning, song2021spatio, 10139790}. These approaches either propose to directly extend the adversarial learning framework~\cite{jamal2018deep} from image-based methods~\cite{ganin2015unsupervised} or couple it with some temporal attention  weights~\cite{chen2019temporal, choi2020shuffle} and self-supervised pretext tasks~\cite{choi2020shuffle, munro2020multi} to align the segment-level features between the domains. However, these strategies produce only a modest $\sim2\%$ gain over the source-only model (see Figure~\ref{fig:teaser}). Recently, there has been a paradigm shift from adversarial to contrastive learning framework~\cite{kim2021learning, song2021spatio, sahoo2021contrast} for video domain adaptation. As shown in Figure~\ref{fig:teaser}, CoMix~\cite{sahoo2021contrast} achieves 6.4\% and 5.1\% gain over the source-only model on the UCF $\rightarrow$ HMDB and HMDB $\rightarrow$ UCF datasets, respectively. However, all these methods are inherently complex and use source domain videos during the adaptation stage, which is untenable in several scenarios~\cite{yang2021exploiting}, as discussed earlier. Due to its practical relevance, source-free domain adaptation is a well-known problem for different computer vision tasks such as image classification~\cite{yang2021exploiting}, and semantic segmentation~\cite{guan2021scale} but relatively under-explored in the context of the video classification task. Therefore, there is a need to investigate source-free domain adaptation for video classification tasks in order to improve the practicality and efficiency of today's approaches.

\begin{figure}[t]
    \centering
    \includegraphics[width=0.7\columnwidth]{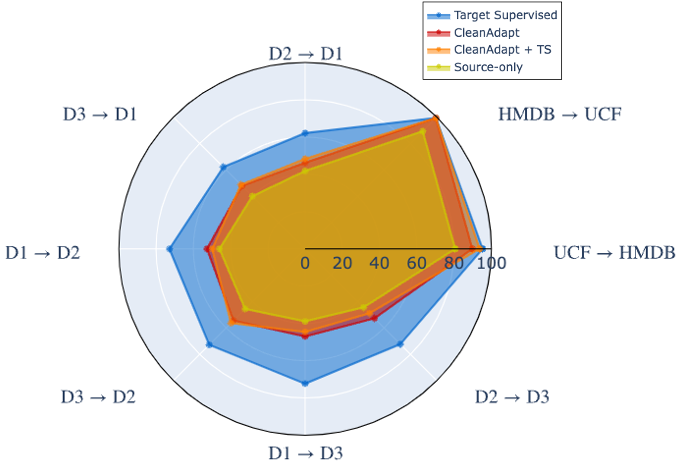}
    \caption{The radar plot illustrates the performance improvements of our proposed methods, \method{} and \methodj{} (shown in \textcolor{orange}{orange} and \textcolor{red}{red}, respectively), compared to the source-only model on multiple benchmarks. The source-only model (shown in \textcolor{yellow}{yellow}), trained on the source domain and tested on the target domain, serves as the lower bound of adaptation performance, while the target-supervised model (shown in \textcolor{blue}{blue}), trained and tested on target domain videos, represents the upper bound. (Best viewed in color.)}
      \label{fig:radar}
\end{figure}

In this work, we present an effective approach that leverages the self-training framework for source-free video UDA, where we do not have access to source-domain videos during the adaptation stage. We generate \textit{pseudo-labels}
for the unlabeled target domain videos using a source pre-trained model. These pseudo-labels are indeed noisy due to the existing domain gap. Finetuning the source pre-trained model with these noisy pseudo-labels is a sub-optimal solution as the presence of incorrect pseudo-labels hinders the adaptation stage, as discussed in Sec.~\ref{sssec:hs}. However, we observe that these pseudo-labeled target domain videos are not completely unusable, and in fact, there is a substantial number of target domain videos with correct pseudo-labels. For example, 
in the case of HMDB $\rightarrow$ UCF, the HMDB pre-trained model produces pseudo-labels with $\sim90\%$ accuracy on the UCF dataset, and we experimentally show that this amount of data is sufficient for adaptation. Throughout this paper, we refer to these samples with correct pseudo-labels as \textit{clean}, whereas the samples with incorrect pseudo-labels are termed \textit{noisy}. 

% We observe that the underlying network learns clean samples first before memorizing the noisy samples, and this acts as the core idea behind the adaptation stage in our proposed method (Figure~\ref{fig:ce}). We discuss this in Sec.~\ref{ss:clean_samples_are_all_you_need}.

\begin{figure*}[ht!]
\centering
        \begin{subfigure}[b]{0.49\textwidth}
        \includegraphics[width=\linewidth]{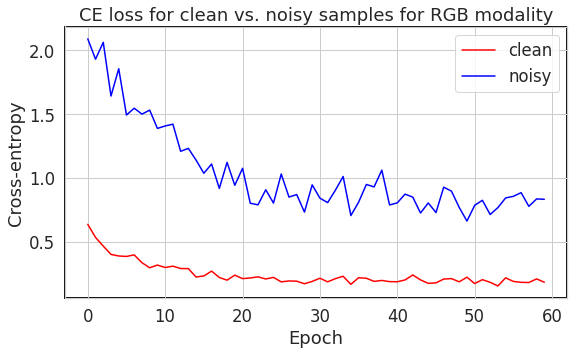}
                \caption{}
                \label{fig:ce_rgb}
        \end{subfigure}%
        \begin{subfigure}[b]{0.49\textwidth}
        \includegraphics[width=\linewidth]{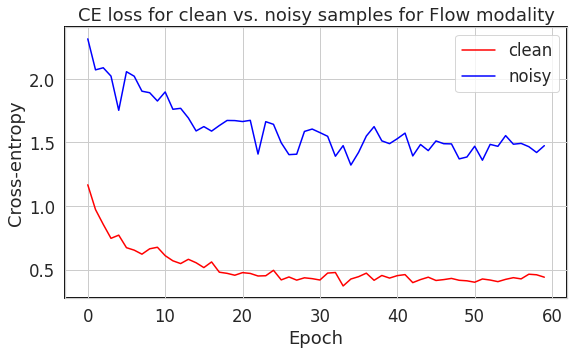}
                \caption{}
                \label{fig:ce_flow}
        \end{subfigure}%
\caption{Average cross-entropy loss per epoch of training with pseudo-labeled target domain videos for clean vs. noisy samples with (a) RGB modality and (b) Flow modality. We term the target domain samples with correct pseudo-labels as \textit{clean} samples and with incorrect pseudo-labels as \textit{noisy} samples. Note that, the groundtruth labels are only used to identify the clean vs. the noisy samples for visualization purposes and not used for training the model.
Deep neural networks learn the clean samples first before memorizing the noisy samples according to the deep memorization effect as proposed in~\cite{arpit2017closer}. In our proposed approach \method{}, we exploit this connection to select the clean samples for fine-tuning the model to adapt to the target domain. (Best viewed in color.)}
\label{fig:ce}
\end{figure*}

We treat the problem of source-free domain adaptation as learning from noisy labels and propose a self-training  based approach that selects clean samples from the noisy pseudo-labeled target domain samples to re-train the model for gradually adapting to the target domain in an iterative manner. We observe that deep neural networks tend to learn from the clean samples first before memorizing the noisy samples in the later stage of training according to the deep memorization effect~\cite{arpit2017closer}. In Figure~\ref{fig:ce}, we validate this effect empirically for both appearance (RGB) and motion (flow) modalities. From Figure~\ref{fig:ce}, it is clear that the model produces small-loss for the clean samples. In contrast to that, the noisy samples have high loss values for both the modalities. In our paper, we exploit this connection between small-loss and clean instances and propose an approach for source-free video domain adaptation. We discuss this in detail in Sec.~\ref{ss:clean_samples_are_all_you_need}.
 This method progressively adapts the model to the target domain in an iterative fashion. Thus, we name our approach as \textit{\method{}}. Additionally, we employ a teacher-student network~\cite{sohn2020fixmatch} to produce more resilient pseudo-labels, where the teacher network is continuously updated by incorporating the temporal ensemble of student networks. This approach generates more consistent pseudo-labels, thus aiding the enhancement of the student network. We refer to this version of our method as \textit{\methodj{}}. Our proposed methods surpass all other source-dependent state-of-the-art methods by a large margin on UCF $\leftrightarrow$ HMDB and EPIC-Kitchens datasets, despite being source-free (see Figure~\ref{fig:teaser} and Figure~\ref{fig:radar}).

An earlier version of our approach was published as a conference paper~\cite{dasgupta2022overcoming}. This manuscript proposes the following significant improvements.
% \textbf{Difference from our conference paper:} This paper is an extension of our conference paper~\cite{dasgupta2022overcoming}. This manuscript significantly improves ~\cite{dasgupta2022overcoming}: 
\begin{enumerate}
    \item We comprehensively survey existing video domain adaptation approaches for different domain adaptation setups in Sec.~\ref{sec:lit}.

    \item We propose a strategy to filter out the incorrect pseudo-labeled (noisy) samples based on the deep memorization effect~\cite{arpit2017closer} and utilize the clean samples in the adaptation stage.
    
    \item In Sec.~\ref{sec:method}, we propose an improved version of our pseudo-label generation process using a teacher-student framework. This modification yields reliable pseudo-label generation, thereby enhancing the overall performance.
    
    \item We conduct a comprehensive analysis of the results obtained with the popular UCF $\leftrightarrow$ HMDB and EPIC-Kitchens datasets and detailed in Sec.~\ref{sec:results}.

    \item We also include evaluations and comparisons in Sec.~\ref{sec:results} with recently developed source-free video domain adaptation techniques like~\cite{xu2022source, yin2022mix, huang2022relative}. Our \method{} and \methodj{} outperform all the existing video domain adaptation approaches.
    
\end{enumerate}

% To our knowledge, we are the first to address the video domain adaptation problem in a source-free setup. 
% We treat this problem as learning from noisy labels and propose a self-training based approach that selects the clean samples from the noisy pseudo-labeled target domain samples to re-train the model for gradually adapting to the target domain in an iterative manner.

% In contrast to the previous methods~\cite{jamal2018deep, chen2019temporal,choi2020unsupervised,pan2020adversarial, choi2020shuffle, munro2020multi,da2022dual, kim2021learning, song2021spatio}, \method{} and \methodj{} are inherently source-free as it only requires target domain videos and their corresponding pseudo-labels. Our proposed method surpasses all other source-dependent video domain adaptation methods by a large margin on UCF $\leftrightarrow$ HMDB and EPIC-Kitchens datasets, despite being source-free.
% Here we extend this work in the following ways: 

The paper is organized as follows. In Section~\ref{sec:lit}, we review some of the notable and recent works in this domain. Section~\ref{sec:method} describes the \method{} and \methodj{} methods. In Section~\ref{sec:results}, we present the experimental analyses and finally conclude in Section~\ref{sec:conclusion}.

%-----------Related Works-----------

\section{Related Work}
\label{sec:lit}
\paragraph*{\bf Supervised action recognition}
Convolutional neural networks (CNNs) are now the de facto solution for action recognition tasks. Various efforts have been made in this context to capture spatio-temporal information in videos, starting from two-stream networks with 2D~\cite{zhou2018temporal} to 3D CNNs~\cite{carreira2017quo}.  
However, a common limitation of existing methods is their dependence on training data that closely matches the distribution of the test data. When there is a subtle difference in the distribution between the training and the test domains, these models struggle to generalize effectively. Consequently, fine-tuning with a large amount of labeled data from the target domain is often required, which can be both time-consuming and expensive. To address this issue, our focus is on unsupervised video domain adaptation, aiming to overcome the need for labeled target domain data.
\begin{table*}[!ht]
\centering
\caption{Summary of domain adaptation for action recognition methods. \textbf{``N/A"} denotes the unavailability of source code.}
\resizebox{\columnwidth}{!}{\begin{tabular}{|llllll|}
\hline
\rowcolor[HTML]{C0C0C0} 
\multicolumn{1}{|c|}{\cellcolor[HTML]{C0C0C0}\textbf{Methods}} & \multicolumn{1}{c|}{\cellcolor[HTML]{C0C0C0}\textbf{Venue}} & \multicolumn{1}{c|}{\cellcolor[HTML]{C0C0C0}\textbf{Code Link}} & \multicolumn{1}{c|}{\cellcolor[HTML]{C0C0C0}\textbf{Backbone}} & \multicolumn{1}{c|}{\cellcolor[HTML]{C0C0C0}\textbf{Core Components}} & \multicolumn{1}{c|}{\cellcolor[HTML]{C0C0C0}\textbf{Types of Videos}} \\ \hline
\hline
%====>

\multicolumn{6}{|c|}{\rule{0pt}{12pt}\textbf{\normalsize{Unsupervised Video Domain Adaptation}}}      \\    \hline
% %======
 \hline
\multicolumn{1}{|c|}{AMLS ~\cite{jamal2018deep}}                        & \multicolumn{1}{c|}{BMVC'18}                                             & \multicolumn{1}{c|}{N/A}                                           & \multicolumn{1}{c|}{C3D}                                          & \multicolumn{1}{c|}{Subspace alignment}                           &           \multicolumn{1}{c|}{Third person}               \\                   
% %======

\multicolumn{1}{|c|}{DAAAA ~\cite{jamal2018deep}}                        & \multicolumn{1}{c|}{BMVC'18}                                             & \multicolumn{1}{c|}{N/A}                                           & \multicolumn{1}{c|}{C3D}                                          & \multicolumn{1}{c|}{Domain invariant feature learning; Adversarial learning with domain discriminator}                           &           \multicolumn{1}{c|}{Third person}               \\                   
% %======

\multicolumn{1}{|c|}{TA3N ~\cite{chen2019temporal}}                        & \multicolumn{1}{c|}{ICCV'19}                                             & \multicolumn{1}{c|}{\href{https://github.com/cmhungsteve/TA3N}{PyTorch}}                          & \multicolumn{1}{c|}{ResNet-101}                                          & \multicolumn{1}{c|}{Attention alignment; Temporal discrepancy; Adversarial alignment}                           &                 \multicolumn{1}{c|}{Third person}    \\
% %======

 \multicolumn{1}{|c|}{TCoN ~\cite{pan2020adversarial}}                        & \multicolumn{1}{c|}{AAAI'20}                                             & \multicolumn{1}{c|}{N/A}                          & 
\multicolumn{1}{c|}{BN-Inception, C3D}                                          & \multicolumn{1}{c|}{Cross-domain co-attention; Adversarial learning for temporal adaptation}                           &                 \multicolumn{1}{c|}{Third person}    \\                   
% %======
 \multicolumn{1}{|c|}{MM-SADA ~\cite{munro2020multi}}                        & \multicolumn{1}{c|}{CVPR'20}                                             & \multicolumn{1}{c|}{\href{https://github.com/jonmun/MM-SADA-code}{Tensorflow}}                          & \multicolumn{1}{c|}{I3D}                                          & \multicolumn{1}{c|}{Self-supervised cross-modal alignment; Adversarial Learning}                           &                 \multicolumn{1}{c|}{First person}    \\

% %======
\multicolumn{1}{|c|}{Choi \etal~\cite{choi2020unsupervised}}                        & \multicolumn{1}{|c|}{WACV'20}                                             & \multicolumn{1}{c|}{N/A}                          & 
\multicolumn{1}{c|}{I3D}                                          & \multicolumn{1}{c|}{Adversarial learning}                           &                 \multicolumn{1}{c|}{Third-person}    \\

% %======
\multicolumn{1}{|c|}{SAVA ~\cite{choi2020shuffle}}                        & \multicolumn{1}{c|}{ECCV'20}                                             & \multicolumn{1}{c|}{N/A}                          & 
\multicolumn{1}{c|}{I3D}                                          & \multicolumn{1}{c|}{Align important clips; Self-supervised clip-order prediction}                           &                 \multicolumn{1}{c|}{Third-person}    \\

% %======
\multicolumn{1}{|c|}{STCDA ~\cite{song2021spatio}}                        & \multicolumn{1}{c|}{CVPR'21}                                             & \multicolumn{1}{c|}{N/A}                          & 
\multicolumn{1}{c|}{BN-Inception, I3D}                                          & \multicolumn{1}{c|}{Spatio-temporal contrastive learning; pseudo-labeling}                           &                 \multicolumn{1}{c|}{First and Third person}    \\
% %======
\multicolumn{1}{|c|}{Kim \etal ~\cite{kim2021domain}}                        & \multicolumn{1}{c|}{ICCV'21}                                             & \multicolumn{1}{c|}{N/A}                          & 
\multicolumn{1}{c|}{I3D}                                          & \multicolumn{1}{c|}{Contrastive learning; cross-modal and cross-domain alignment}                           &                 \multicolumn{1}{c|}{First and Third person}    \\
% %======
\multicolumn{1}{|c|}{CoMix ~\cite{sahoo2021contrast}}                        & \multicolumn{1}{c|}{NeurIPS'21}                                             & \multicolumn{1}{c|}{\href{https://github.com/CVIR/CoMix}{PyTorch}}                             & 
\multicolumn{1}{c|}{I3D}                                          & \multicolumn{1}{c|}{Temporal contrastive learning; background mixing; pseudo-labeling}                           &                 \multicolumn{1}{c|}{First and Third person}    \\

% %======
% \multicolumn{1}{|c|}{MAN ~\cite{gao2021novel}}                        & \multicolumn{1}{c|}{TC'21}                                             & \multicolumn{1}{c|}{N/A}                             & 
% \multicolumn{1}{c|}{ResNet-152}                      &               \multicolumn{1}{c|}{Multiple-view adversarial
% learning network}                      & 
% \multicolumn{1}{c|}{Third person}    \\
% %======
\multicolumn{1}{|c|}{CO2A ~\cite{da2022dual}}                        & \multicolumn{1}{c|}{WACV'22}                                             & \multicolumn{1}{c|}{\href{https:
//github.com/vturrisi/CO2A.}{PyTorch}}                             & 
\multicolumn{1}{c|}{I3D}                                          & \multicolumn{1}{c|}{Dual head contrastive network; Synthetic data}                           &                 \multicolumn{1}{c|}{Third person}    \\

% %======
\multicolumn{1}{|c|}{M$A\textsuperscript{2}$LTD ~\cite{chen2022multi}}                        & \multicolumn{1}{c|}{WACV'22}                                             & \multicolumn{1}{c|}{\href{https://github.com/justchenpp/MA2L-TD}{PyTorch}}                             & 
\multicolumn{1}{c|}{ResNet-101}                                          & \multicolumn{1}{c|}{Multi-level temporal features; Multiple domain discriminators}                           &                 \multicolumn{1}{c|}{Third person}    \\
% %======
\multicolumn{1}{|c|}{CIA ~\cite{yang2022interact}}                        & \multicolumn{1}{c|}{CVPR'22}                                             & \multicolumn{1}{c|}{N/A}                             & 
\multicolumn{1}{c|}{I3D}                                          & \multicolumn{1}{c|}{Cross-modal complementarity and consensus}                           &                 \multicolumn{1}{c|}{First and Third person}    \\
% %======
% \multicolumn{1}{|c|}{ACAN ~\cite{xu2022aligning}}                        & \multicolumn{1}{c|}{TNNLS'22}                                             & \multicolumn{1}{c|}{N/A}                             & 
% \multicolumn{1}{c|}{I3D, MFNet}                      &               \multicolumn{1}{c|}{Pixel correlation alignment}                      & \multicolumn{1}{c|}{Third person}    \\

% %======
\multicolumn{1}{|c|}{TranSVAE ~\cite{wei2022unsupervised}}                        & \multicolumn{1}{c|}{NeurIPS'23}                                             & \multicolumn{1}{c|}{\href{https://github.com/ldkong1205/TranSVAE}{PyTorch}}                                 & 
\multicolumn{1}{c|}{I3D}                      &               \multicolumn{1}{c|}{Disentanglement of domain-related and semantic-related information}                      & 
\multicolumn{1}{c|}{First and Third person}    \\
% %======
% \multicolumn{1}{|c|}{Wu \etal ~\cite{wu2022dynamic}}                        & \multicolumn{1}{c|}{Neurocomputing'22}                                             & \multicolumn{1}{c|}{N/A}                                 & 
% \multicolumn{1}{c|}{ResNet-50}                      &               \multicolumn{1}{c|}{Video-level mix-up learning; Dynamic sampling strategy}                      & 
% \multicolumn{1}{c|}{Third person}    \\
% %======
% \multicolumn{1}{|c|}{Zhang \etal ~\cite{zhang2022audio}}                        & \multicolumn{1}{c|}{CVPR'22}                                             & \multicolumn{1}{c|}{\href{https://github.com/xiaobai1217/DomainAdaptation}{PyTorch}}    
% & 
% \multicolumn{1}{c|}{ConvLSTM, I3D, Timesformer}                      &               \multicolumn{1}{c|}{Frame dependency modeling}                      & 
% \multicolumn{1}{c|}{Third person}    \\ 
% %======
\multicolumn{1}{|c|}{MD-DMD \etal ~\cite{yin2022mix}}                        & \multicolumn{1}{c|}{MM'22}                                             & \multicolumn{1}{c|}{N/A}                                 & 
\multicolumn{1}{c|}{I3D}                      &               \multicolumn{1}{c|}{Dynamic modal distillation}                      & 
\multicolumn{1}{c|}{First and Third person}    \\ 
% %======
\multicolumn{1}{|c|}{Broome \etal ~\cite{broome2023recur}}                        & \multicolumn{1}{c|}{WACV'23}                                             & \multicolumn{1}{c|}{N/A}                                 & 
\multicolumn{1}{c|}{SlowFast (video), Resnet-18 (audio)}                      &               \multicolumn{1}{c|}{Audio-adaptive encoder }                      & 
\multicolumn{1}{c|}{First and Third person}    \\ 

% %======
\multicolumn{1}{|c|}{CTAN~\cite{10139790}}                        & \multicolumn{1}{c|}{TCSVT'23}                                             & \multicolumn{1}{c|}{\href{https://github.com/xianyuanliu/CTAN}{PyTorch}}                                 & 
\multicolumn{1}{c|}{I3D}                      &               \multicolumn{1}{c|}{Channel-temporal attention network}                      & 
\multicolumn{1}{c|}{First person}    \\ 
\hline \hline

% %====================================>
\multicolumn{6}{|c|}{\textbf{\rule{0pt}{12pt}\normalsize{Source-free Video Domain Adaptation}}}      \\    \hline\hline
% %======
\multicolumn{1}{|c|}{\method{}~\cite{dasgupta2022overcoming}}                        & \multicolumn{1}{c|}{ICVGIP'22}                                             & \multicolumn{1}{c|}{\href{https://github.com/avijit9/CleanAdapt}{PyTorch}}                                 & 
\multicolumn{1}{c|}{I3D}                      &               \multicolumn{1}{c|}{Pseudo-labeling; Learning from noisy labels}                      & 
\multicolumn{1}{c|}{First and Third person}    \\ 
% %======
\multicolumn{1}{|c|}{ATCoN~\cite{xu2022source}}                        & \multicolumn{1}{c|}{ECCV'22}                                             & \multicolumn{1}{c|}{\href{https://github.com/xuyu0010/ATCoN}{PyTorch}}                                 & 
\multicolumn{1}{c|}{ResNet-50}                      &               \multicolumn{1}{c|}{Temporal consistency network}                      & 
\multicolumn{1}{c|}{Third person}    \\ 
% %======
\multicolumn{1}{|c|}{MTRAN ~\cite{huang2022relative}}                        & \multicolumn{1}{c|}{MM'22}                                             & \multicolumn{1}{c|}{N/A}                                 & 
\multicolumn{1}{c|}{I3D, Transformer}                      &               \multicolumn{1}{c|}{Temporal relative alignment; Mix-up}                      & 
\multicolumn{1}{c|}{First and Third person}    \\ 
\hline %\hline

\end{tabular}}
\label{tab:lit}
\end{table*}

\paragraph*{\bf Domain adaptation for action recognition}
Early works~\cite{chen2019temporal,munro2020multi, choi2020shuffle} on video UDA are inspired by the adversarial framework~\cite{ganin2015unsupervised} for image-based UDA tasks. Jamal \etal~\cite{jamal2018deep} proposes to align the source and the target domains using a subspace alignment technique and outperform all the previous shallow methods. Chen \etal~\cite{chen2019temporal} show the efficacy of attending to the temporal dynamics of video for domain adaptation. TCoN~\cite{pan2020adversarial} is a cross-domain co-attention module for matching the source and the target domain features with appearance and motion streams. Munro \etal~\cite{munro2020multi} were among the first to show the effectiveness of learning multi-modal correspondence for video domain adaptation. SAVA~\cite{choi2020shuffle} is an attention-augmented model with a clip order prediction task to re-validate the effectiveness of self-supervised learning for video domain adaptation, as shown in~\cite{munro2020multi}. Overall, the adversarial methods are complex and sensitive to the choice of hyperparameters~\cite{sahoo2021contrast}. 

There has been a recent shift from adversarial to contrastive learning-based methods for the video UDA task. Song~\etal~\cite{song2021spatio} propose to bridge the domain gap using a self-supervised contrastive framework named cross-modal alignment. In a similar direction, Kim \etal~\cite{kim2021learning} use a cross-modal feature alignment loss for learning a domain adaptive feature representation. CoMix~\cite{sahoo2021contrast} represents videos as graphs and uses temporal-contrastive learning over graph representations for transferable feature learning. Additionally, these methods~\cite{song2021spatio, kim2021learning, sahoo2021contrast} generate pseudo-labels from the source pre-trained model for the target domain videos and use only the target domain samples with high-confident pseudo-labels in their contrastive loss in each iteration. However, the source-only model often makes incorrect predictions with high confidence due to the distribution shift for target domain videos, which can hinder adaptation. To address this, we treat target pseudo-labels as noisy and formulate the domain adaptation problem as learning from noisy labels. Moreover, the adaptation stage in these methods~\cite{song2021spatio, kim2021learning, sahoo2021contrast} is \textit{source-dependent}. This is an impractical requirement as source data transfer during the deployment phase of the model is often infeasible.

Recently, ATCoN~\cite{xu2022source} and our conference paper CleanAdapt~\cite{dasgupta2022overcoming} have addressed this issue of source data dependency. These methods introduce a source-free adaptation approach, \ie, it does not rely on source domain videos during the adaptation stage. In Table \ref{tab:lit}, we provide an overview of existing methods for unsupervised and source-free video domain adaptation. Xu et al.~\cite{xu2022video} provide an extensive survey of video domain adaptation, encompassing a variety of setups.
%Other works ~\cite{chen2021conditional,wang2022calibrating,xu2021partial} have made notable advancements in tackling various domain adaptation scenarios for videos, including open-set, partial-set, and test-time setups. 

\paragraph*{\bf Learning from noisy-labels}
Self-training based methods with careful design choices may still produce over-confident, incorrect predictions. To alleviate this issue, we resort to learning from label-noise literature. One of the popular approaches to reducing the effect of noisy-labels is to design noise-robust losses~\cite{feng2021can}. However, these methods fail to handle real-world noise~\cite{zhang2021prototypical}. 
According to~\cite{arpit2017closer}, deep neural networks produce small loss values for samples with correct pseudo-labels. Thus, a popular direction for handling label-noise is to use the cross-entropy loss to indicate label correctness~\cite{han2018co} and leverage these small-loss samples for re-training the networks. In this work, we demonstrate that the small-loss samples are potentially clean samples and are effective in helping our source pre-trained model adapt to the target domain if these samples are used for fine-tuning. Therefore, our proposed approach is simpler that the existing approaches, requiring solely pseudo-labeled samples from the target domain.

%-----------Methodology-----------
\section{Approach}
\label{sec:method}

\begin{figure*}[ht]
    \centering
    \includegraphics[width=\textwidth]{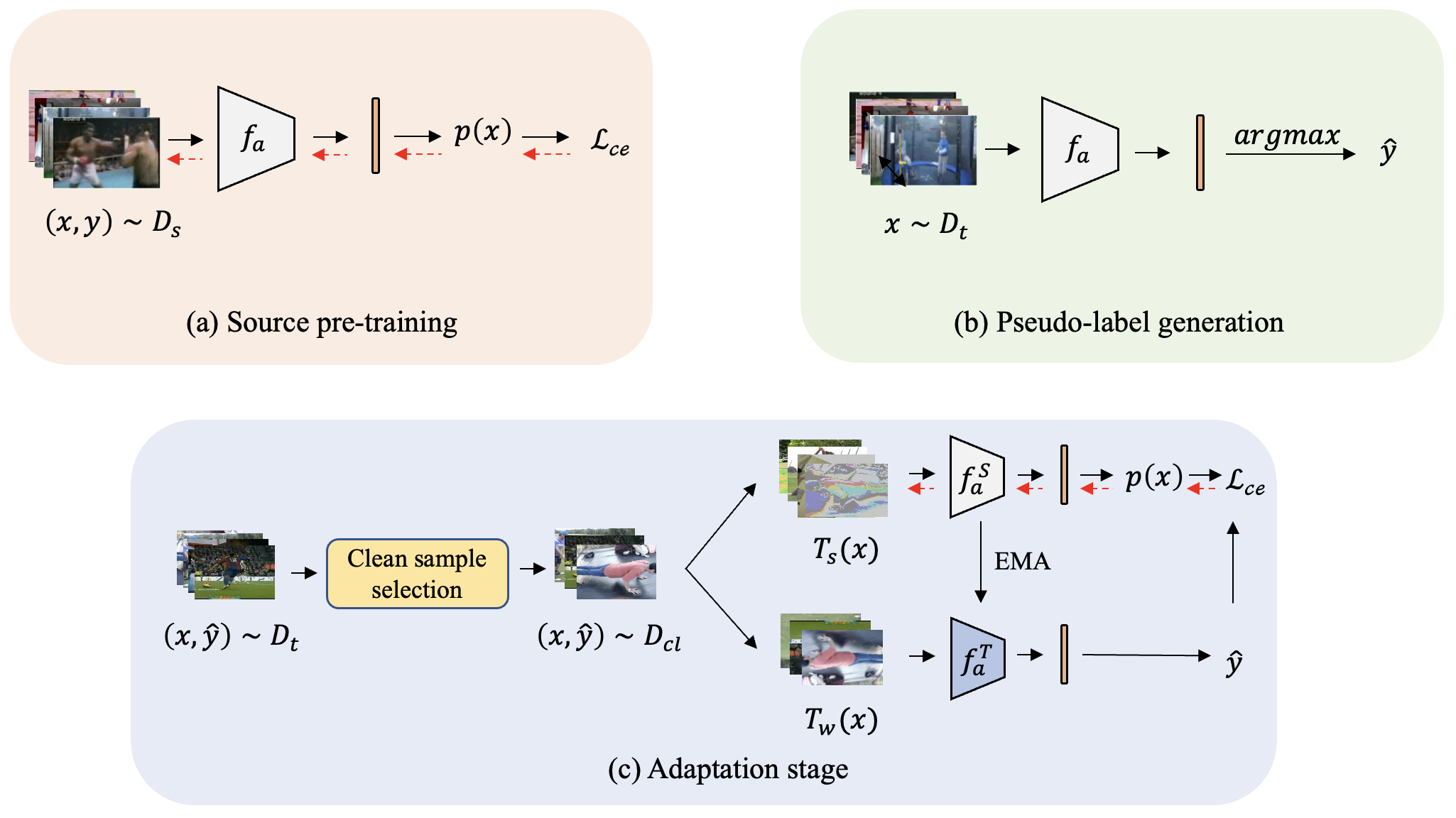}
    \caption{Overview of the three stages of our \methodj{} framework for source-free video domain adaptation, which has three stages. (a) The model ($f_a$) is first pre-trained on the \textit{labeled} source domain videos from $\mathcal{D}_s$. For brevity, only the single-stream model is shown here. (b) This source pre-trained model is then used to generate pseudo-labels $\hat{y}$ for the \textit{unlabeled} target domain videos from $\mathcal{D}_t$. Inevitably, these pseudo-labels are noisy due to the domain shift between the source and the target domains. (c) A \textit{clean sample selection} module is used to select a set $\mathcal{D}_{cl}$ of small-loss samples as potential clean samples. The source pre-trained model is finetuned on these clean samples from $\mathcal{D}_{cl}$ using their corresponding pseudo-labels $\hat{y}$. We repeat this step multiple times. See Sec.~\ref{sec:results} for implementation details. (Best viewed in color.)}
    \label{fig:pipeline}
\end{figure*}
\subsection{Problem Definition}
 In the \textit{source-free} UDA task for videos, we are given a labeled source domain dataset of videos $D_s = \{(x_s, y_s): x_s \sim P\}$, where $P$ is the source data distribution and $y_s$ is the corresponding label of $x_s$. We are also given an unlabeled target domain dataset $D_t = \{x_t : x_t \sim Q\}$, where $Q$ is the target distribution that is different from the source distribution $P$. We assume that the source and the target domains share the same label-set $C$, \ie, closed-set domain setup. 

For a video clip $x$ from any domain, we consider two modalities, $x = \{x_a, x_m\}$, where $x_a$ is the appearance (RGB) stream and $x_m$ is the motion (optical flow) stream. We use two 3D CNN backbones $f_a$  and $f_m$, one for each modality that classifies a video into one of the $|C|$ classes. We aim to adapt the 3D CNNs ($f_a$ and $f_m$) to the target domain. We also note that the source domain videos are only available during the pre-training stage and we do not use this dataset $\mathcal{D}_s$ during the adaptation stage as we are interested in the more realistic source-free setup. We show an overview of the proposed method in Figure~\ref{fig:pipeline}.

% \begin{figure}[t]
%     \centering
%     \includegraphics[width=\columnwidth]{assets/ce-loss.png}
%     \caption{Plot of average cross-entropy loss per epoch of training with pseudo-labeled target domain videos for clean vs. noisy samples with RGB and flow modalities. 
%     % We term the target domain samples with correct pseudo-labels as \textit{clean} and with incorrect pseudo-labels as \textit{noisy}.
%     Note that, the groundtruth labels are only used to identify the clean vs. the noisy samples for visualization purpose and not used for training the model. 
%     %(Best viewed in color.) }
%     % Deep neural networks learn the clean samples first before memorizing the noisy samples according to the deep memorization effect as proposed in~\cite{arpit2017closer}. In our proposed approach \method{}, we exploit this connection to select the clean samples for fine-tuning the model to adapt to the target domain.
%     (Best viewed in color.)}
%      \label{fig:ce}
% \end{figure}

\begin{figure}[t]
    \centering
    \includegraphics[width=0.7\columnwidth]{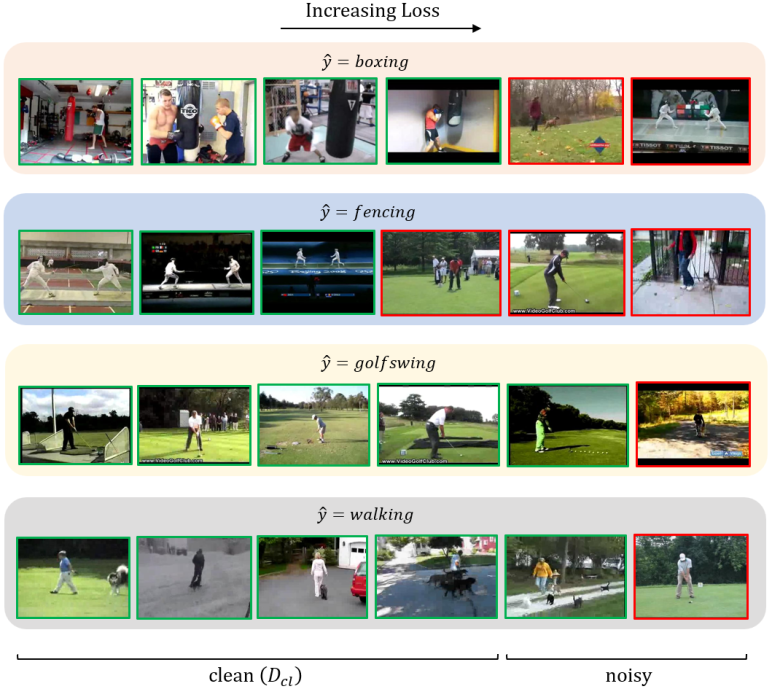}
    \caption{The \textit{clean sample selection} module. The pseudo-labeled target domain videos from $\mathcal{D}_t$ are grouped according to their pseudo-labels $\hat{y}$ and sorted in ascending order of the loss generated by the model against their pseudo-labels. The \textit{keep-rate} $\tau$ ($\tau = 0.6$ in this example) decides the number of samples to be selected for adaptation, having small-loss values for each class. For simplicity, we have used only four classes here. We show the videos with the correct pseudo-labels inside \textcolor{green}{green} border, whereas the videos with incorrect pseudo-labels are inside the \textcolor{red}{red} border solely for visualization purposes. (Best viewed in color.)}
    \label{fig:slt}
\end{figure}

\subsection{Self-training based Domain Adaptation}
In contrast to the adversarial learning based approaches~\cite{munro2020multi, chen2019temporal, choi2020shuffle}, we take the path of self-training primarily due to its simplicity in the adaptation stage. First, we pre-train the 3D CNN models using the labeled source videos from $D_s$. Second, we generate pseudo-labels for the unlabeled target dataset $D_t$ using the source pre-trained model referred to as pseudo-labels. Third, we retrain the networks $f_a$ and $f_m$ using the pseudo-labeled target domain videos from $D_t$ for adaptation. One of the possibilities is to use all the samples with their corresponding pseudo-labels to retrain the networks. However, pseudo-labels are noisy due to the domain gap between the source domain $D_s$ and the target domain $D_t$. Retraining $f_a$ and $f_m$ with all these pseudo-labeled samples from $D_t$ leads to a sub-optimal result, as discussed in Section~\ref{sec:results}. We aim to answer the following question in this paper: How do we choose the pseudo-labeled samples from $D_t$ that can help the model in adaptation?

\subsection{Clean Samples are All You Need}
\label{ss:clean_samples_are_all_you_need}
The pseudo-labels contain a large number of samples with correct pseudo-labels (clean samples). For example, there are $\sim$90\% samples with correct pseudo-labels in the UCF dataset when generated using the HMDB pre-trained networks. Thus, if we can filter out the noisy samples and keep only the clean ones, we can finetune our networks ($f_a$ and $f_m$) using these clean samples and their corresponding pseudo-labels. Thus, we argue that these clean samples are the ones that can help us in domain adaptation. Now, the important question here is how to separate the clean samples from the noisy ones.

% To this end, we cast the problem of video domain adaptation as learning from noisy-labels due to noisy pseudo-labels. 

In Figure~\ref{fig:ce}, we observe that deep neural networks learn from clean samples easily and have a hard time learning from noisy samples due to the memorization effect~\cite{arpit2017closer}. Thus, samples with \textit{low-loss} values are the potential clean samples and can be filtered out using the loss as an indicator. In this work, we design two approaches without bells and whistles, \textit{\method} and \textit{\methodj}, aiming to select the clean samples based on the loss generated by the model against their corresponding pseudo-labels for adaptation. In each epoch of the adaptation stage, we select these clean samples from the target domain and use them to re-train the source pre-trained models $f_a$ and $f_m$.

There are three key advantages to this: (1) we do not need to modify the overall training regime (\eg, contrastive learning for domain alignment~\cite{kim2021learning, song2021spatio,sahoo2021contrast}) during adaptation, (2) we do not need to make any domain adaptation specific design choices (\eg, background mixing~\cite{sahoo2021contrast}), and (3) we implicitly design an adaptation method that \textit{does not} need any source data during the adaptation stage (refer Figure~\ref{fig:pipeline}).

\subsection{Source Pre-training}
\label{sub:source-only}

In the source pre-training stage, we train the 3D CNNs $f_a$ and $f_m$ using the labeled source domain dataset $D_s$, and we refer to these as \textit{source-only} models. For a sample $(x, y) \in D_s$, we average the logits obtained from $f_a(x)$ and $f_m(x)$ to compute the final score $p(x)$ as follows:  

\begin{equation}
p(x) = \sigma(f_a(x) + f_m(x)).
\end{equation}
We use the conventional cross-entropy loss between the predicted class probabilities $p(x)$ and the one-hot encoded ground-truth label $y$ as the loss function for training:
\begin{equation}
\mathcal{L}_{ce}(x) = - \sum_{c = 1}^{|C|} y^c \log(p^c(x)),
\label{eq:ce}
\end{equation}
where $y^c$ and $p^c$ represent the $c^{th}$ element of $y$ and $p(x)$ respectively for class $c$. The main goal for this pre-training step is to equip our model with the initial knowledge of the classes present in the source dataset $D_s$. Figure~\ref{fig:pipeline}(a) depicts this step.

\subsection{Pseudo-label Generation}
\label{ss:plg}
The next step, as illustrated in Figure~\ref{fig:pipeline}(b), is to generate the pseudo-labels for the unlabeled target domain samples.
Once the model is pre-trained on the source domain videos, we use the learned notion of the class semantics of the model to generate labels for the target domain data. Note that these generated labels are not the actual labels for the target domain videos. Thus, we term these source-only model-generated labels as \textit{pseudo-labels} $\hat{y}$. Formally,
\begin{equation}
\hat{y}(x) = \argmaxA_c p^c(x),
\end{equation}
where $x \in D_t$. Due to the domain shift between the source and the target, these pseudo-labels $\hat{y}$ are noisy.

\subsection{\methodj{}: A Strong Video Adaptation Method}

Once the pseudo-labels are obtained for the target domain videos, we use them for adaptation, as shown in Figure~\ref{fig:pipeline}(c). As discussed earlier, the pseudo-labels are noisy, and we aim to extract samples with the correct pseudo-labels (clean samples) for adaptation. Each epoch of the adaptation stage has two key steps in our framework: (a) clean sample selection and (b) fine-tuning the models $f_a$ and $f_m$ using these clean samples.

\textbf{Clean sample selection.} To filter out the target domain videos with noisy pseudo-labels, we start with the pseudo-labels generated in Sec.~\ref{ss:plg} and exploit the relation between the small-loss and the clean samples. We use the source pre-trained models ($f_a$, $f_m$) to select the clean samples reliably.
In each epoch, the videos are first grouped into $|C|$ classes based on their pseudo-labels generated by the model and sorted in ascending order of their cross-entropy loss values computed using the \textit{prediction} made by the models ($p(x) = \sigma(f_a(x) +f_m(x))$) and their corresponding \textit{pseudo-labels} ($\hat{y}$):

\begin{equation}
    D_{cl}, D_{no} \xleftarrow{} \mathcal{L}_{ce}(p(x), \hat{y}(x)),
\end{equation}
where $\sigma(.)$, $D_{cl}$, and $D_{no}$ represent the softmax function, sets of clean samples and noisy samples, respectively. If the pseudo-labels are correct, the model is likely to produce a small loss, and thus, there is a high possibility that the sample belongs to $D_{cl}$. Inspired by~\cite{han2018co}, we define a hyper-parameter \textit{keep-rate} $\tau$. For each group, we select $\tau$ proportion of the total number of samples with small losses (See Figure~\ref{fig:slt}). We call this updated dataset of small-loss samples as $D_{cl} \subset D_{t}$ and discard the samples in $D_{no}$. We update the pseudo-labels as follows: 
\begin{equation}
    \hat{y}(x) \xleftarrow{} \argmaxA_c p^{c}(x).
\end{equation}
This version of our proposed model is referred to as \method.
%This step is illustrated in Figure~\ref{fig:slt}.

Inspired by the success of the teacher-student framework~\cite{tarvainen2017mean, sohn2020fixmatch}, we adopt it along with our small-loss based clean sample selection to combat the label-noise, and we call this model as \textit{\methodj}. The teacher and student networks share the same architecture and are initialized with the source pre-trained weights. We use temporally averaged teacher models to select reliable pseudo-labeled target domain videos. To accomplish this, we generate two copies of the models $f_a$ and $f_m$: one operates as the teacher model ($f_a^T$, $f_m^T$), while the other functions as the student model ($f_a^S$, $f_m^S$) in their respective modalities. The parameters of the teacher models ($\theta_a^T$ and $\theta_m^T$) are not updated through loss back-propagation, rather they are an exponential moving average of student parameters $\theta_a^S$ and $\theta_m^S$ respectively. To this end, we create two different versions of the same video $x$ in each modality using two transformations. Let $T_w(x)$ and $T_s(x)$ denote the weakly and strongly augmented versions of the same video $x \in \mathcal{D}_t$ as defined below:

\textbf{Weak augmentations.} We refer to common geometric transformations like flipping and shifting as weak augmentations. In particular, we incorporate a random horizontal flip, applied universally to the video, with a 50\% probability.

\textbf{Strong augmentations.} To achieve robust augmentation, we implement RandAugment~\cite{cubuk2020randaugment} for each video $x$. This method randomly selects three augmentations from a predefined list and applies them to the video $x$.

The teacher networks, denoted as $f_a^T$ and $f_m^T$, use the weakly augmented version of the video $x$ to produce pseudo-labels. In contrast, the student networks, represented by $f_a^S$ and $f_m^S$, undergo fine-tuning using the strongly augmented version of the video $x$. This fine-tuning process enhances their resilience to noise.

As for \method{}, here also the samples are divided into the clean and the noisy sets as follows: 

\begin{equation}
    D_{cl}, D_{no} \xleftarrow{} \mathcal{L}_{ce}(p^T(x), \hat{y}(x)),
\end{equation}

where $p^T(x) = \sigma(f^T_a(T_w(x)) +f^T_m(T_w(x)))$. We also update the pseudo-labels using the teacher model, as shown below.
\begin{equation}
    \hat{y}(x) \xleftarrow{} \argmaxA_c p^{T, c}(T_w(x)),
\end{equation}
where $P^{T,c}$ denotes the probability of the sample $x$ belonging to class $c$ as predicted by the teacher model.

% To generate pseudo-labels, we feed the weakly augmented video $T_w(x)$ to the teacher networks $f_a^T(x)$ and $f_m^T(x)$ and generate the 
% \textit{pseudo-labels} $\hat{y}$. Formally,
% \begin{equation}
% \hat{y} = \argmaxA_c p^c(T_w(x)),
% \end{equation}
% where $x \in D_t$. Due to the domain shift between the source and the target, these pseudo-labels $\hat{y}$ are noisy. 

\textbf{Fine-tuning.}  In this step, the student networks $f^S_a$ and $f^S_m$ are re-trained using the strongly augmented samples $T_s(x)$ and their corresponding pseudo-label $\hat{y(x)}$ from $D_{cl}$ using the cross-entropy loss as shown below.
\begin{equation}
\mathcal{L}_{ce}(x) = - \sum_{c = 1}^{|C|}\hat{y}^c \log(p^{S, c}(T_s(x))).
\label{eq:ce-finetune}
\end{equation}
where $(x, \hat{y}) \in D_{cl}$ and $P^{S,c}$ denotes the probability of the sample $x$ belonging to class $c$ as determined by the student model. The parameters of the teacher networks are updated using the exponential moving average of the updated student networks as follows:

\begin{equation}
\begin{split}
    & \theta_a^T \xleftarrow{} \epsilon \theta_a^T + (1 - \epsilon) \theta_a^S, \\
    &\theta_m^T \xleftarrow{} \epsilon \theta_m^T + (1 - \epsilon) \theta_m^S,
    \end{split}
\end{equation}
where $\epsilon$ is the momentum parameter. We repeat these two steps in an iterative manner until the networks converge.

%-----------Results-----------
\section{Results and Analysis}
\label{sec:results}

\subsection{Datasets and Metrics} 
We consider both first-person and third-person videos for benchmarking our proposed approach. Following~\cite{song2021spatio, sahoo2021contrast}, we use publicly available UCF101~\cite{soomro2012ucf101} and HMDB51~\cite{kuehne2011hmdb} for third-person and EPIC-Kitchens~\cite{damen2018scaling} for first-person videos. We show experimentally that our approach adapts well to both of these scenarios.

\textbf{UCF  $\leftrightarrow$ HMDB.} We use the official split released by Chen \etal~\cite{chen2019temporal} for UCF $\leftrightarrow$ HMDB to evaluate our \method{} on video domain adaptation. In total, this dataset has 3209 third-person videos with 12 action classes. Specifically, all the videos are a subset of the original UCF101~\cite{soomro2012ucf101} and HMDB51~\cite{kuehne2011hmdb} datasets with 12 classes common between them. Following~\cite{chen2019temporal}, we use two settings: UCF101 $\rightarrow$ HMDB51, and HMDB51 $\rightarrow$ UCF101. 

\textbf{UCF  $\leftrightarrow$ HMDB\textsubscript{small}.} This dataset has 5 shared classes from UCF101 and HMDB51 datasets with a total of 1271 videos. 

\textbf{EPIC-Kitchens.} This is the largest video domain adaptation dataset, which contains egocentric videos of fine-grained actions recorded in different kitchens. We follow the official split provided by Munro \etal~\cite{munro2020multi}. This dataset contains videos from the three largest kitchens, \ie, D1, D2, and D3, with 8 common action categories. EPIC-Kitchens has more class imbalance than UCF $\leftrightarrow$ HMDB, making it more challenging~\cite{munro2020multi}.

\textbf{Ego2Exo.} We examine and evaluate the effectiveness of our proposed approach for cross-view transfer in videos using the challenging Ego2Exo~\cite{kalluri2024lagtran} dataset. This dataset includes videos sourced from the Ego-Exo4D~\cite{grauman2024ego} dataset, leveraging their keystep annotations for action labels (\eg, \textit{make dough}, \textit{prepare skillet}, \etc). In total, it comprises 4,100 ego videos and 4,986 exo videos for training, while the validation set consists of 3,168 samples from each view.

\textbf{Metrics.} We follow the standard protocol defined by~\cite{chen2019temporal, munro2020multi} to compare our approach with state-of-the-art unsupervised domain adaptation methods~\cite{kim2021learning, song2021spatio, sahoo2021contrast} in terms of top-1 accuracy. We perform cross-domain retrieval experiments to evaluate the feature space learned by our model before and after adaptation. We report retrieval performance in terms of Recall at $k$ (R@k), implying that if $k$ closest nearest neighbors contain one video of the same class, the retrieval is considered correct.

\subsection{Implementation Details}

We use the Inception I3D~\cite{carreira2017quo} network as the backbone for both RGB and Flow modalities. Following the prior video domain adaptation works~\cite{munro2020multi,choi2020shuffle, kim2021learning, song2021spatio}, we use the Kinetics~\cite{carreira2017quo} pre-trained weights to initialize the I3D network. During training, we randomly sample 16 consecutive frames and perform the same data augmentation used in~\cite{munro2020multi,choi2020shuffle,kim2021learning} for all our steps. We set the batch size to 48 for both UCF $\leftrightarrow$ HMDB and EPIC-Kitchens datasets. We pre-compute optical flow using the TV-L1 algorithm.

\textbf{Source pretraining stage.} We train the model on the source dataset for 40 and 100 epochs with learning rates $1e-2$, and  $2e-2$ for UCF $\leftrightarrow$ HMDB and EPIC-Kitchens datasets, respectively. We reduce the learning rate by a factor of 10 after 10, 20 epochs for UCF $\leftrightarrow$ HMDB. For EPIC-Kitchens, we decrease the learning rate by 10 after 50 epochs. We follow~\cite{choi2020shuffle} for other hyperparameters.

\setlength{\tabcolsep}{4pt}
\begin{table*}[h!]
\centering
\caption{Performance comparison with state-of-the-art video domain adaptation methods on UCF101$\leftrightarrow$ HMDB51. Result for MM-SADA~\cite{munro2020multi} is taken from Kim \etal~\cite{kim2021learning}. The results for our methods are highlighted in \colorbox{lightgray}{gray}.
The average gains over the source-only model for the first and second best source-free approaches are highlighted in \textcolor{ForestGreen}{green} and \textcolor{BrickRed}{red}, respectively.
}
% \begin{adjustbox}{max width=0.7\textwidth}
\resizebox{0.7\textwidth}{!}{
\begin{tabular}{lccccccclclc}

\hline
\multirow{2}{*}{\textbf{Method}} & \multirow{2}{*}{\textbf{Venue}}&\multirow{2}{*}{} & \multirow{2}{*}{\textbf{Two-stream?}}& \multirow{2}{*}{} & \multirow{2}{*}{\textbf{Source-free?}} &\multirow{2}{*}{\textbf{Backbone}} &  & \multicolumn{4}{c}{\textbf{Datasets}} \\ \cline{8-11} 
                  &                   &                   &                   &                   &  &&&   \textbf{UCF $\rightarrow$ HMDB}  &     &   \textbf{HMDB $\rightarrow$ UCF}  &    \\ \hline

Source only~\cite{chen2019temporal} & & & & && I3D & & 80.6 &  & 88.8 &     \\ 
TA3N~\cite{chen2019temporal}&ICCV'19& & \textcolor{red}{\xmark} &   & \textcolor{red}{\xmark} & I3D&&81.4    &&90.5  &    \\
Target supervised~\cite{chen2019temporal}&&  &  && &I3D &  &  93.1  & &  97.0  &  \\ \hline

Source only~\cite{choi2020shuffle} & & & & &     &I3D&& 80.3   &  &   88.8 &  \\ 
SAVA~\cite{choi2020shuffle}&ECCV'20& & \textcolor{red}{\xmark} & &\textcolor{red}{\xmark}& I3D   & & 82.2   & &     91.2  &  \\
Target supervised~\cite{choi2020shuffle}&& & & & & I3D  &   & 95.0     & &     96.8  &      \\ \hline

Source only~\cite{munro2020multi} & &&& &  &I3D &   &    82.8 &&  90.7    &   \\ 
MM-SADA~\cite{munro2020multi}&CVPR'20&& \textcolor{blue}{\cmark}  && \textcolor{red}{\xmark}    &I3D&&   84.2    &&  91.1  &   \\
Target supervised~\cite{munro2020multi} & &&& &  &I3D & &98.8    & &   95.0   &   \\ \hline  

Source only~\cite{song2021spatio} & &&& &  &I3D &   & 82.8    &&     89.8  &   \\ 
STCDA~\cite{song2021spatio}&CVPR'21&& \textcolor{blue}{\cmark}  && \textcolor{red}{\xmark}    &I3D&&  83.1     &&     92.1 &   \\
Target supervised~\cite{song2021spatio} & &&& &  &I3D & & 95.8      & &     97.7 &   \\ \hline  

Source only~\cite{kim2021learning} & &&& &  &I3D &   &    82.8 &&  90.7    &   \\ 
Kim \etal~\cite{kim2021learning}&ICCV'21&& \textcolor{blue}{\cmark}  && \textcolor{red}{\xmark}    &I3D&&   84.7    &&  92.8  &   \\
Target supervised~\cite{kim2021learning} & &&& &  &I3D & &98.8    & &   95.0   &   \\ \hline  

Source only~\cite{sahoo2021contrast} & &&& &  &I3D &   &    80.3 &&  88.8    &   \\ 
CoMix~\cite{sahoo2021contrast}&NeurIPS'21&& \textcolor{red}{\xmark}  && \textcolor{red}{\xmark}    &I3D&&  86.7    &&  93.9  &   \\
Target supervised~\cite{sahoo2021contrast} & &&& &  &I3D & &95.0    & &   96.8   &   \\ \hline  
%Wu \etal~\cite{wei2022unsupervised}&Neurocomputing'21&& \textcolor{red}{\xmark}  && \textcolor{red}{\xmark}& ResNet-50    &&  86.9     &&    98.6  &  \\  \hline
Source only~\cite{xu2022source} & &&& &  &TRN &   & 72.8  && 72.2  &   \\
ATCoN~\cite{xu2022source}&ECCV'22&& \textcolor{red}{\xmark}  && \textcolor{blue}{\cmark}    &TRN&& 79.7 \textcolor{BrickRed}{$\blacktriangle$ +6.9} && 85.3 \textcolor{ForestGreen}{$\blacktriangle$ +13.1}  &   \\ \hline

Source only~\cite{yin2022mix} & &&& &  &I3D &   &  80.8   &&   91.0 &   \\ 
MD-DMD~\cite{yin2022mix}&MM'22&& \textcolor{blue}{\cmark}  && \textcolor{red}{\xmark}    &I3D&& 82.2  &&  92.8  &   \\
Target supervised~\cite{yin2022mix} & &&& &  &I3D & &  98.8  & &  95.0 &   \\ \hline  

Source only~\cite{huang2022relative} & &&& &  &Transformer &   & 81.1  && 86.8  &   \\
MTRAN~\cite{huang2022relative}&MM'22&& \textcolor{blue}{\cmark}  && \textcolor{blue}{\cmark}    &Transformer&& 92.2 \textcolor{ForestGreen}{$\blacktriangle$ +11.1}   && 95.3 \textcolor{BrickRed}{$\blacktriangle$ +8.5} &   \\ \hline

% Source only~\cite{kim2021learning} & & &  & & & 82.8     &&     90.7 &   \\ 
% MM-SADA~\cite{munro2020multi}&& \textcolor{blue}{\cmark}  && \textcolor{red}{\xmark}    &&  84.2      &&     91.1  \\
% Kim \etal~\cite{kim2021learning}&& \textcolor{blue}{\cmark}  && \textcolor{red}{\xmark}    &&  84.7  & &        92.8  &  \\
% Target supervised~\cite{kim2021learning} & &&&   &   & 98.8    &&     95.0  &\\ \hline  

% Source only~\cite{sahoo2021contrast} &  & &&&    & 80.3      &&     88.8 &   \\ 
% CoMix~\cite{sahoo2021contrast}&& \textcolor{red}{\xmark}  && \textcolor{red}{\xmark}    &&  86.7     &&     93.9  &  \\
% Target supervised~\cite{sahoo2021contrast} &  &&&   &  & 95.0   &&      96.8  &   \\ \hline  

Costa \etal~\cite{da2022dual}&WACV'22&& \textcolor{red}{\xmark}  && \textcolor{red}{\xmark}&I3D    &&  87.8     &&     95.8  &  \\  \hline

Source only &  & &  &&&ResNet-101&  & 76.4   &&     78.1 &     \\ 
M$A\textsuperscript{2}$LTD ~\cite{chen2022multi} &WACV'22&& \textcolor{red}{\xmark}  && \textcolor{red}{\xmark}&ResNet-101    &&   85.0    &&   86.6   &  \\  \hline

Source only~\cite{yang2022interact} & &&& &  &I3D &   &  85.8   &&   93.5 &   \\ 
CIA~\cite{yang2022interact}&CVPR'22&& \textcolor{blue}{\cmark}  && \textcolor{red}{\xmark}    &I3D-TRN && 91.9  && 94.6  &   \\
Target supervised~\cite{yang2022interact} & &&& &  &I3D & &  96.8  & &  99.1 &   \\ \hline 

Source only~\cite{wei2022unsupervised} & &&& &  &I3D &   &  86.1   &&   92.5 &   \\ 
TranSVAE~\cite{wei2022unsupervised}&NeurIPS'23&& \textcolor{blue}{\cmark}  && \textcolor{red}{\xmark}    &I3D-TRN && 87.8  && 99.0  &   \\
 \hline

% Source only~\cite{xu2022aligning} & &&& &  &I3D &   &  80.3   &&  88.8 &   \\ 
% ACAN~\cite{xu2022aligning}&TNNLS'22&& \textcolor{red}{\xmark} && \textcolor{red}{\xmark}    &I3D-TRN && 85.4  && 91.2  &   \\
% Target supervised~\cite{xu2022aligning} & &&& &  &I3D & &  95.0  & &  96.8 &   \\ \hline 

% % % %% RGB
\rowcolor{Gray}
Source only &  & &  &&&I3D&  & 80.6   &&     89.3 &     \\ \rowcolor{Gray}

\method{}&Ours&& \textcolor{red}{\xmark}  && \textcolor{blue}{\cmark}    &I3D&& 86.1 \textcolor{black}{$\blacktriangle$ +5.5}  &&  96.1 \textcolor{black}{$\blacktriangle$ +6.8} & \\
\rowcolor{Gray}
\methodj{}&Ours&& \textcolor{red}{\xmark}  &&  \textcolor{blue}{\cmark}    &I3D&& 88.6 \textcolor{black}{$\blacktriangle$ +8.0}   && 96.7                
\textcolor{black}{$\blacktriangle$ +7.4} & \\
\rowcolor{Gray}
Target supervised & &&& &  &I3D & &93.6    & &   98.4   &   \\ \hline  

\rowcolor{Gray}
Source only &  & &  &&&I3D&  &82.5  &&   91.4 &     \\ \rowcolor{Gray}
\method{}&Ours&& \textcolor{blue}{\cmark}  &&  \textcolor{blue}{\cmark}    &I3D&& 89.8 \textcolor{black}{$\blacktriangle$ +7.3}   && 99.2 \textcolor{black}{$\blacktriangle$ +7.8} & \\
\rowcolor{Gray}
\methodj{}&Ours&& \textcolor{blue}{\cmark}  &&  \textcolor{blue}{\cmark}    &I3D&& 93.6 \textcolor{ForestGreen}{$\blacktriangle$ +11.1}   && 99.3 \textcolor{black}{$\blacktriangle$ +7.9} & \\
\rowcolor{Gray}
Target supervised & &&& &  &I3D & &95.3    & &   99.3   &   \\ \hline  

\end{tabular}
}
% \end{adjustbox}
\label{tab:ucf-hmdb}
\end{table*}
\textbf{Adaptation stage.} We use the
source pre-trained weights during the adaptation stage to initialize the I3D~\cite{carreira2017quo} network. The network is trained for 100 epochs with learning rates $1e-2$ and $2e-3$ for UCF $\leftrightarrow$ HMDB and EPIC-Kitchens, respectively. The learning rate is reduced by 10 after 20, 40 epochs for UCF $\leftrightarrow$ HMDB. In the case of EPIC-Kitchens, we reduce the learning rate by 10 after 10, 20 for EPIC-Kitchens. We set the values of momentum parameters $\epsilon$ as 0.99 in all experiments.

Our entire framework is implemented in PyTorch and uses 4 NVIDIA 2080Ti GPUs. On average, training takes around 1 hour for UCF $\leftrightarrow$ HMDB and approximately 7 hours for EPIC-Kitchens datasets.

We now first provide a detailed comparison of our proposed approaches with state-of-the-art video domain adaptation methods on UCF  $\leftrightarrow$ HMDB, UCF  $\leftrightarrow$ HMDB\textsubscript{small}, and EPIC-Kitchens datasets in Sec.~\ref{sec:comparison-with-sota}. We provide some discussions to understand the effect of high-loss samples and over-confident pseudo-labels on the adaptation stage. In Sec.~\ref{sec:comparison-with-sota}, we also show and compare the heatmaps generated by our \method{} and source pre-trained model. In Sec.~\ref{sssec:hs}, we illustrate the impact of the hyperparameter $\tau$ and explore considerations for selecting an appropriate value for it. We also experimentally show the retrieval performance of our proposed approach as well as the source pre-trained model in Sec.~\ref{sec:retrieval}.
In Sec.~\ref{sec:ts}, we discuss the impact of the teacher-student framework in detail.

\subsection{Comparisons to the State-of-the-art Methods}
\label{sec:comparison-with-sota}

\textbf{UCF $\leftrightarrow$ HMDB.} We present the quantitative results of both of our approaches \method{} and \methodj{} for UCF $\leftrightarrow$ HMDB dataset in Table~\ref{tab:ucf-hmdb} and compare our results with the state-of-the-art unsupervised source-free video domain adaptation approaches. For each approach in Table~\ref{tab:ucf-hmdb}, we also report \textit{source-only} and \textit{target-supervised} results for fair comparisons wherever applicable. The source-only method refers to the $f_a$ and/or $f_m$ models trained only on the {\fontfamily{qcr}\selectfont train} split of the source dataset as described in Section~\ref{sub:source-only} and tested directly on the {\fontfamily{qcr}\selectfont validation} split of the target dataset, which serves as a lower bound of the adaptation performance. The target-supervised model is trained and tested on the {\fontfamily{qcr}\selectfont train} and {\fontfamily{qcr}\selectfont validation} split of the target dataset, respectively. This serves as an upper bound to the adaptation performance.

Next, we shift our focus towards comparing our method with the most advanced unsupervised video domain adaptation techniques available.
TA3N~\cite{chen2019temporal}, SAVA~\cite{choi2020shuffle}, CoMix~\cite{sahoo2021contrast}, ATCoN~\cite{xu2022source}, M$A\textsuperscript{2}$LTD~\cite{chen2022multi} and Costa \etal~\cite{da2022dual} use only appearance stream in their methods. In contrast to these methods, STCDA~\cite{song2021spatio}, MM-SADA~\cite{munro2020multi}, and Kim \etal~\cite{kim2021learning}, MD-DMD~\cite{yin2022mix}, MTRAN~\cite{huang2022relative}, CIA~\cite{yang2022interact}, and TransVAE~\cite{wei2022unsupervised} leverage both appearance and motion streams. We show the results for both single-stream and two-stream versions of our model. 

To show that the efficacy of our proposed approach is not solely due to the addition of the motion stream with appearance, we show our adaptation results for both single-stream (appearance only) and two-stream (appearance and motion) models. Our single-stream model achieves \textbf{86.1\%} and \textbf{96.1\%} top-1 accuracy with a gain of \textbf{5.5\%} and \textbf{6.8\%} over the source-only model for UCF $\rightarrow$ HMDB and HMDB $\rightarrow$ UCF datasets respectively. Further improvement in the adaptation performance is observed when we couple our method \method{} with the teacher-student framework~\cite{tarvainen2017mean, sohn2020fixmatch} resulting in \methodj{}. The teacher network serves as a regularization mechanism by producing consistent pseudo-labels, which, in turn, incentivizes the student model to make more confident predictions. This improved method \methodj{} achieves \textbf{88.6\%} and \textbf{96.7\%} top-1 accuracy, resulting in \textbf{8.0\%}
and \textbf{7.4\%} gains over the source-only models, respectively.

In comparison, the best performing earlier existing model CoMix~\cite{sahoo2021contrast}, which uses a temporal contrastive learning framework with background mixing, gives 6.4\% and 5.1\% gain for these two datasets, respectively. Note that all of these methods use the source data along with the target data during adaptation, whereas we use only target data in our approach and attain similar gains.
Although the source-free video domain adaptation approaches such as ATCoN~\cite{xu2022source} and MTRAN~\cite{huang2022relative} achieve better performance for HMDB $\xrightarrow{}$ UCF than our proposed approaches \method{} and \methodj{}, it is not consistent across all the datasets as shown in Table~\ref{tab:ucf-hmdb} and Table~\ref{tab:epic}.

Similarly, our two-stream model \method{} achieves state-of-the-art performance on both UCF $\rightarrow$ HMDB and HMDB $\rightarrow$ UCF datasets in terms of top-1 accuracy with the values of \textbf{89.8\%}
and \textbf{99.2\%}, respectively. This is a significant gain of \textbf{7.3\%} for UCF $\rightarrow$ HMDB and \textbf{7.8\%} for HMDB $\rightarrow$ UCF over the source-only model without using any source-domain data which is much higher than the other source-dependent adaptation models. Our improved method \methodj{} achieves a further gain of \textbf{11.1\%} for UCF $\rightarrow$ HMDB and \textbf{7.9\%}. This affirms the assertion that in source-free, unsupervised video domain adaptation, utilizing the low-loss samples from the target domain during the adaptation phase is justified. It also highlights the efficacy of employing a slowly updated teacher network for generating pseudo-labels to fine-tune the student network using strongly augmented target domain videos.

We now aim to show the effect of using high-loss samples for adaptation and discuss if overconfident pseudo-labels can affect adaptation performance. 

\textbf{What happens if we use only high-loss samples for adaptation?} We trained our two-stream network with the high-loss samples instead of the proposed low-loss samples. For UCF $\rightarrow$
HMDB, we obtained 84.7\% accuracy after adaptation
with the high-loss samples, which is 5.1\% less when adapted with the low-loss samples. We observe a similar drop for HMDB $\rightarrow$ UCF. This difference
is even more significant when the noisy pseudo-labels
are dominant (e.g., more than 12\% on Epic-Kitchens). Nevertheless, these outcomes are in line with expectations, as demonstrated in Figure~\ref{fig:ce}, where it is evident that the noisy samples typically exhibit elevated loss values, thereby detrimentally affecting the fine-tuning performance when incorporated during the adaptation stage.
\begin{table}[ht!]
\addtolength{\tabcolsep}{-0.3em}
\centering
\begin{adjustbox}{max width=\textwidth}
\begin{minipage}{0.48\textwidth}
\footnotesize
    \centering
    \caption{Comparison with state-of-the-art image-based source-free domain adaptation techniques.}
    \begin{tabular}{lccccccc}
    \hline
    {\textbf{Method}} & \textbf{Backbone} & &  &  \textbf{U $\rightarrow$ H} &  & \textbf{H $\rightarrow$ U}  &  \\ \hline
    Source only & TRN & & & 72.7 & & 72.2& \\
    Kim \etal~\cite{kim2021domain} & TRN & & & 69.9 & & 74.9  & \\
    Li \etal~\cite{li2020model} & TRN & & & 74.4 & & 67.3  & \\
    Yang \etal~\cite{yang2020unsupervised} & TRN & & & 75.3 & & 76.3  & \\
    Qiu \etal~\cite{qiu2021source}& TRN & & & 75.8 & & 68.2  & \\\hline
    Source only & I3D & & & 80.6 & & 89.3 & \\
    Yang \etal~\cite{yang2021exploiting} & I3D & & & 86.6 & & 91.4 & \\
    Liang \etal~\cite{liang2020we} & I3D & & & 82.5 & & 91.9 & \\
    \method{} & I3D & & & 86.1 & & 96.1 & \\
    \methodj{} & I3D & & & \textbf{88.6} & & \textbf{96.7} & \\\hline
    \label{tab:source-free}
    \end{tabular}
\end{minipage}
% \hfill
\begin{minipage}{0.48\textwidth}
\footnotesize
    \centering
    \caption{Performance comparisons with state-of-the-art video domain adaptation methods on UCF  $\leftrightarrow$ HMDB\textsubscript{small}.}
    \begin{tabular}{lccccccc}
    \hline
    {\textbf{Method}} &  &  &  \textbf{U $\rightarrow$ H} &  & \textbf{H $\rightarrow$ U}  &  \\ \hline
    Source only  & & & 97.3 & & 96.8& \\
    SHOT~\cite{liang2020we} & & & 99.3 & & 99.5& \\
    3C-GAN~\cite{li2020model} & & & 98.3 & & 99.5& \\
    SFDA~\cite{kim2021domain} & & & 98.0 & & 99.3& \\
    HCL~\cite{huang2021model}& & & 99.3 & & 99.5& \\
    MTRAN~\cite{huang2022relative} & & & 100 & & 100& \\
    \hline
    Source only  & & & 98.3 & & 97.8& \\
    \methodj{} & & & 100 & & 100& \\
    Target Supervised  & & & 100 & & 100  &\\
    \hline
    \label{tab:uh-small}
    \end{tabular}
\end{minipage}
\end{adjustbox}
\end{table}

\setlength{\tabcolsep}{4pt}
\begin{table*}[h!]
% \scriptsize
\centering
\caption{Performance comparison with state-of-the-art video domain adaptation methods on EPIC-Kitchens dataset. Results with single-stream models are highlighted in \colorbox{lightcyan}{cyan} whereas the results with two-stream networks are highlighted in \colorbox{lightgray}{gray}. The average gains over the source-only model for the first and second best source-free approaches are highlighted in \textcolor{ForestGreen}{green} and \textcolor{BrickRed}{red}, respectively.
}
\resizebox{\textwidth}{!}{
\begin{tabular}{cccclllllll}
\hline
 \textbf{Method} & \textbf{Venue} & \textbf{Source-free?} & \textbf{Backbone}  &\textbf{D2$\rightarrow$D1} & \textbf{D3$\rightarrow$D1} & \textbf{D1$\rightarrow$D2} & \textbf{D3$\rightarrow$D2} & \textbf{D1$\rightarrow$D3} & \textbf{D2$\rightarrow$D3} & \textbf{Mean}   \\ \hline 
  \rowcolor{Gray}
 Source only & && I3D &42.5 &  44.3 & 42.0 & 56.3 & 41.2 & 46.5 &  45.5  \\  \rowcolor{Gray}
 MMD~\cite{long2015learning}& ICML'15& \textcolor{red}{\xmark} &I3D &43.1 &  48.3 & 46.6& 55.2  & 39.2 & 48.5 & 46.8\\ \rowcolor{Gray}
 AdaBN~\cite{li2018adaptive}&PR'18 & \textcolor{red}{\xmark} & I3D &44.6 & 47.8  & 47.0 & 54.7 & 40.3 & 48.8 &  47.2\\ \rowcolor{Gray}
 MCD~\cite{saito2018maximum}& CVPR'18 & \textcolor{red}{\xmark} & I3D &42.1 & 47.9  &  46.5 & 52.7 & 43.5  & 51.0 & 47.3  \\ \rowcolor{Gray}
 MM-SADA~\cite{munro2020multi}&CVPR'20& \textcolor{red}{\xmark} & I3D &48.2 & 50.9  & 49.5 & 56.1 & 44.1 & 52.7 &  50.3 \\ \rowcolor{Gray}
 STCDA~\cite{song2021spatio}&CVPR'21 & \textcolor{red}{\xmark} & I3D &49.0  & 52.6 & 52.0 & 55.6 & 45.5 & 52.5 & 51.2  \\ \rowcolor{Gray}
 Kim \etal~\cite{kim2021learning}&ICCV'21& \textcolor{red}{\xmark} & I3D &49.5 & 51.5 & 50.3 & 56.3 & 46.3 & 52.0 & 51.0  \\ \rowcolor{Gray}
 MD-DMD~\cite{yin2022mix}&MM'22 & \textcolor{red}{\xmark} & I3D & 50.3 & 51.0& 56.0 & 54.7& 47.3&52.4 &52.0  \\\rowcolor{Gray}
CIA~\cite{yang2022interact}&CVPR'22 & \textcolor{red}{\xmark} & I3D & 52.5 & 47.8& 49.8 & 53.2& 52.2&57.6&52.2 \\ \rowcolor{Gray}
 Target Supervised & && I3D &62.8 & 62.8 & 71.7 & 71.7 & 74.0 & 74.0 &   69.5 \\
 \hline

\rowcolor{LightCyan}
 Source only & & & I3D&35.5 & 38.1 & 39.4 & 40.5 & 32.0 & 39.2 & 37.5 \\ \rowcolor{LightCyan}
CTAN~\cite{10139790}&TCSVT'23 &\textcolor{red}{\xmark}& I3D & 36.6 &39.3& 41.3&41.3 &35.0  &40.6 &39.0\\ 
 \rowcolor{LightCyan}
 Target Supervised &&&I3D & 60.2 & 60.2 & 64.7  & 64.7 & 52.8  & 52.8  &  59.2 \\ \hline

 \rowcolor{LightCyan}
 Source only & &&I3D&  35.4 &  34.6 &32.8  & 35.8&34.1  &39.1  & 35.3 \\ 
 \rowcolor{LightCyan}
 DANN~\cite{ganin2015unsupervised} & ICML'15&\textcolor{red}{\xmark}&I3D& 38.3 & 38.8 & 37.7 & 42.1&36.6 &41.9 &39.2\\ 
 \rowcolor{LightCyan}
ADDA~\cite{tzeng2017adversarial}& CVPR'17 &\textcolor{red}{\xmark} &I3D &36.3  & 36.1 & 35.4 & 41.4& 34.9& 40.8&37.4 \\
 TA3N~\cite{chen2019temporal}& ICCV'19 &\textcolor{red}{\xmark} &I3D& 40.9& 39.9&34.2 &44.2 &37.4 &42.8 & 39.9\\ 
 \rowcolor{LightCyan}
CoMix~\cite{sahoo2021contrast} & NeurIPS'21 &\textcolor{red}{\xmark} &I3D& 38.6 &42.3 & 42.9 &49.2 &40.9 &45.2 &43.2 \\ 
 \rowcolor{LightCyan}
 Target Supervised&&&I3D& 57.0 & 57.0 & 64.0 & 64.0 & 63.7 & 63.7   & 61.5\\
 \hline
  \rowcolor{Gray}
 Source only & && Transformer & 43.7 & 51.1 & 40.5 &36.2 &48.9 & 45.2 &  44.2 \\ 
 \rowcolor{Gray}
Liang \etal~\cite{liang2020we} &ICML'20& \textcolor{blue}{\cmark}& Transformer &  44.1 & 53.9 & 40.8 & 36.5 & 49.0 & 45.3 & 44.9 \\ 
  \rowcolor{Gray}
Li \etal~\cite{li2020model} &CVPR'20& \textcolor{blue}{\cmark}& Transformer &44.7   &54.3 &41.0  & 36.7 & 49.9& 45.4  & 45.4 \\ 
  \rowcolor{Gray}
Kim \etal~\cite{kim2021domain} &TAI'21& \textcolor{blue}{\cmark}& Transformer & 44.4 & 54.9 & 41.3 & 37.2 & 49.8 & 45.2  & 45.5 \\  
 \rowcolor{Gray}
HCL~\cite{huang2021model} &NeurIPS'21& \textcolor{blue}{\cmark}& Transformer & 45.1 &55.6&41.5 &36.9 & 50.2 &  45.7 &45.8  \\  
 \rowcolor{Gray}
MTRAN~\cite{huang2022relative} &MM'22 &\textcolor{blue}{\cmark}& Transformer & 46.3 &58.2& 42.2&38.1 &52.3  &46.1  &47.2 \textcolor{BrickRed}{$\blacktriangle$ +3.0}  \\  \hline

\rowcolor{LightCyan}
 Source only & & & I3D&40.9 & 38.6 & 39.3 & 41.3 & 37.3 & 42.4 & 39.9 \\ \rowcolor{LightCyan}
 \method{} & Ours &\textcolor{blue}{\cmark} &I3D&  44.6 \textcolor{black}{$\blacktriangle$ +3.7}& 40.7 \textcolor{black}{$\blacktriangle$ +2.1 } &  44.5 \textcolor{black}{$\blacktriangle$ +5.2} &  47.1 \textcolor{black}{$\blacktriangle$ +5.8 }&  40.9 \textcolor{black}{$\blacktriangle$ +3.6} & 45.7 \textcolor{black}{$\blacktriangle$ +3.3}&  43.9 \textcolor{black}{$\blacktriangle$ +4.0}  \\ \rowcolor{LightCyan}
 Target Supervised &&&I3D & 60.5 & 60.5 & 68.4  & 68.4 & 68.8  & 68.8  &  65.9 \\ \hline

 \rowcolor{Gray}
 Source only && &I3D&41.8 & 40.0 & 46.0  & 45.6 & 38.9 & 44.4 &  42.8  \\ \rowcolor{Gray}
 \method{} &Ours& \textcolor{blue}{\cmark} &I3D& 46.2 \textcolor{black}{$\blacktriangle$ +4.4}& 47.8 \textcolor{black}{$\blacktriangle$ +7.8} & 52.7 \textcolor{black}{$\blacktriangle$ +6.7} & 54.4 \textcolor{black}{$\blacktriangle$ +8.8}& 47.0 \textcolor{black}{$\blacktriangle$ +8.1} &  52.7 \textcolor{black}{$\blacktriangle$ +8.3}&  50.3 \textcolor{ForestGreen}{$\blacktriangle$ +7.5}  \\ \rowcolor{Gray}
 Target Supervised && & I3D&62.1 & 62.1 & 72.8  & 72.8  & 72.3 & 72.3  & 69.1 \\ \hline

 \rowcolor{Gray}
 Source only && &I3D&41.8 & 41.1 & 41.9  & 46.1 & 37.3 & 43.9 &  42.0  \\ \rowcolor{Gray}
 \methodj{} &Ours& \textcolor{blue}{\cmark} &I3D& 48.3\textcolor{black}{$\blacktriangle$ +6.5}& 48.7 \textcolor{black}{$\blacktriangle$ +7.6} & 49.9 \textcolor{black}{$\blacktriangle$ +8.0} & 56.3 \textcolor{black}{$\blacktriangle$ +10.2}& 44.6 \textcolor{black}{$\blacktriangle$ +7.3} &  48.9 \textcolor{black}{$\blacktriangle$ +5.0}&  49.6 \textcolor{ForestGreen}{$\blacktriangle$ +7.6}  \\ \rowcolor{Gray}
 Target Supervised && & I3D&62.3 & 62.3 & 72.7  & 72.7  & 71.1 & 71.1  & 68.4 \\ \hline
\end{tabular}}
\label{tab:epic}
\end{table*}

\textbf{Do the overconfident pseudo-labels trigger error accumulation?} 
A potential question arising here is whether the degradation of models and the further decline in the quality of pseudo-labels can occur due to the presence of \textit{overconfident yet incorrect} pseudo-labels. Although error accumulation could be a possibility,
we have found error accumulation to be negligible in practice.
For example, the UCF pre-trained model selects low-loss samples with $\sim$98\% accuracy in each epoch of the
adaptation stage from HMDB.

\textbf{Comparisons with self-training based methods.} In Table~\ref{tab:ucf-hmdb}, we compare our approach with the other self-training approaches~\cite{kim2021learning, sahoo2021contrast, song2021spatio}. Our method re-purposed the learning from noisy labels based pseudo-label selection method that shows better performance than all
these.

\textbf{Comparisons with image-based source-free methods.} In Table~\ref{tab:source-free}, we compare our approach with state-of-the-art image-based source-free methods. For~\cite{kim2021domain, li2020model, yang2020unsupervised, qiu2021source}, we report the values with TRN~\cite{zhou2017temporalrelation} as their backbone network. Our model \methodj{} achieves 
higher gain over their corresponding source-only model than all these image-based source-free methods. We
have also adopted the frameworks proposed by~\cite{liang2020we, yang2021exploiting} with our 3D backbone network. Liang \etal~\cite{liang2020we} perform marginally better than the source-only model. Yang \etal~\cite{yang2021exploiting} performance is comparable to ours on UCF $\rightarrow$ HMDB, but significantly worse on HMDB$\rightarrow$UCF.

\textbf{UCF  $\leftrightarrow$ HMDB\textsubscript{small}.} To provide a thorough evaluation and compare fairly with existing methods, we compare the performance of our method \methodj{} on the UCF  $\leftrightarrow$ HMDB\textsubscript{small} dataset in Table ~\ref{tab:uh-small}. The dataset size is very limited, and thus \methodj{} achieves scores comparable to the target-supervised baseline for both the UCF  $\rightarrow$ HMDB\textsubscript{small} and HMDB $\leftrightarrow$ UCF\textsubscript{small}. We compare our approach \methodj{} with state-of-the-art techniques such as ~\cite{liang2020we, kim2021domain, huang2022relative, li2020model, huang2021model} and achieves superior performance.

\textbf{EPIC-Kitchens.} In Table~\ref{tab:epic}, we compare the results of our approach with state-of-the-art image-based methods extended for videos as well as video-based domain adaptation. We implement our model to replicate the source-only and target-supervised performance as reported in~\cite{kim2021learning}. Note that there is a minor difference ($\sim$2.7\% and $\sim$3.5\%) in the performance of the source-only model reported in MM-SADA~\cite{munro2020multi} and both of our models. Comparable distinctions to those observed in~\cite{munro2020multi} can be identified in~\cite{sahoo2021contrast}, primarily attributable to the non-deterministic operations associated with CUDA. However, such minor differences of source-only
accuracy is not a concern for evaluating domain adaptation
performance. The most important metric here is the
gain achieved after adaptation over the source-only model. On average, the source-free methods demonstrate a maximum improvement of 3\% over the source-only model. In contrast, our improved method, \methodj{}, achieves 7.6\% improvement despite being simple.
% \begin{figure*}[h!]
%     \centering
%     \includegraphics[width=0.8\textwidth]{assets/cleanadapt vs cleanadapt_ts.png}
%     \caption{Analysis of the effect of \methodj{} on the target validation accuracy as compared to \method{}.(Best viewed in color.)}
%       \label{fig:ts_effect}
% \end{figure*}

% \begin{figure}[h!]
%     \centering
%     \includegraphics[width=0.8\columnwidth]{assets/uh-ablation.png}
%     \caption{Hyperparameter search for the value of keep-rate $\tau$ for UCF101$\leftrightarrow$ HMDB51. The keep-rate $\tau$ controls the number of samples to be selected as clean having low-loss values computed against the pseudo-labels generated by the source-only model. All results reported here are for two-stream network. See supplementary for EPIC-Kitchens. (Best viewed in color.)} 
    
%     \label{fig:split-ratio}
% \end{figure}

% \begin{figure*}[h!]
%     \centering
%     \includegraphics[width=0.8\textwidth]{assets/retrieval-main-paper.png}
%     \caption{Nearest neighbour retrieval results for the UCF $\rightarrow$ HMDB dataset. The left column shows the query videos from the target domain. The middle column shows the retrieved source videos using the source-only model, and the right column shows the source videos retrieved using our proposed model. (Best viewed in color.)}
    
%     \label{fig:retrieval}
% \end{figure*}

MM-SADA~\cite{munro2020multi} is the first to report domain adaptation results on the EPIC-Kitchens dataset, achieving an average of 4.8\% gain on top of their source-only model followed by Song \etal~\cite{song2021spatio} reporting an average gain of 5.7\%. Kim \etal~\cite{kim2021learning} show an improvement of 5.5\% averaged over 6 datasets. CTAN~\cite{10139790} achieves a modest gain of 1.5\% over the source-only models. 
However, all of these methods use the source dataset for adaptation. In contrast to these prior approaches, our simple yet powerful source-free approaches, \method{} and \methodj{}, achieve an average of \textbf{7.5\%} and \textbf{7.6\%} gain over the source-only model, respectively. The primary source of the performance boost achieved by our methods, despite their simplicity, can be attributed to the abundance of clean samples with low-loss values compared to the noisy ones.

\begin{table}[htbp]
\centering
\footnotesize
\caption{Performance comparison of state-of-the-art video domain adaptation method on Ego2Exo~\cite{kalluri2024lagtran} dataset.}
\label{tab:ego-exo}
\begin{tabular}{lccc}
\toprule
\textbf{} & \textbf{Ego$\rightarrow$Exo} & \textbf{Exo$\rightarrow$Ego} & \textbf{Avg.} \\
\midrule
\multicolumn{4}{l}{\textbf{Unsupervised Adaptation}} \\
Source Only & 8.39 & 15.66 & 12.03 \\
TA3N~\cite{chen2019temporal} & 6.92 & 27.95 & 17.44 \\
TransVAE~\cite{wei2022unsupervised} & 12.06 & 23.34 & 17.70 \\
\midrule
\multicolumn{4}{l}{\textbf{Zero-shot Video Recognition}} \\
EgoVLP~\cite{lin2022egocentric} & 5.89 & 19.35 & 12.62 \\
LaViLA~\cite{zhao2023learning} & 5.86 & 19.13 & 12.50 \\
% TextMatch & 10.36 & 13.57 & 11.97 \\
% nGramMatch & 11.50 & 14.56 & 13.98 \\
\midrule
LaGTran~\cite{kalluri2024lagtran} & 12.34 & 30.76 & 21.55 \\
\midrule
Target Sup. & 17.91 & 33.19 & 25.55 \\
\midrule
\midrule
\multicolumn{4}{l}{\textbf{Source-free Unsupervised Adaptation}} \\
Source-only & 18.08 & 40.02 & 29.05 \\
CleanAdapt ($\tau = 0.5$) & 27.14 & 50.47 & 38.81 \\
Target Sup. & 30.99 & 52.05 & 41.52 \\
\bottomrule
\end{tabular}
\end{table}

\textbf{Ego2Exo.} Following~\cite{kalluri2024lagtran}, we use the pre-extracted Omnivore Swin-L~\cite{girdhar2022omnivore} features in the Ego-Exo4D~\cite{grauman2024ego} dataset. We train a 2-layer linear classifier on top of this features keeping all other hyper-parameters as described above.
%We examine and evaluate the effectiveness of our proposed approach for cross-view transfer in videos using the challenging Ego2Exo~\cite{kalluri2024lagtran} dataset. This dataset includes videos sourced from the Ego-Exo4D~\cite{grauman2024ego} dataset, leveraging their keystep annotations for action labels (\eg, \textit{make dough}, \textit{prepare skillet}, \etc). In total, it comprises 4,100 ego videos and 4,986 exo videos for training, while the validation set consists of 3,168 samples from each view.}

% \begin{figure}[h]
% \begin{subfigure}{.99\textwidth}
%   \centering
%   % include first image
%   \includegraphics[width=0.9\linewidth]{assets/uh-ablation-ego2exo.png} 
%   \label{fig:ucf-hmdb}
% \end{subfigure}
% \caption{Effect of keep-rate $\tau$ on Ego2Exo~\cite{kalluri2024lagtran} dataset. (Best viewed in color.)}
% \label{fig:split-ratio-ego-exo}
% \end{figure}

As shown in Table~\ref{tab:ego-exo}, TA3N~\cite{chen2019temporal} and TransVAE~\cite{wei2022unsupervised} achieve average improvements of 5.41\% and 5.67\%, respectively, over the source-only model on the Ego2Exo dataset. In contrast, the zero-shot baselines, EgoVLP~\cite{lin2022video} and LaViLA~\cite{zhao2023learning}, deliver more modest gains of 0.59\% and 0.47\%, respectively. Although LaGTran~\cite{kalluri2024lagtran} achieves a 9.52\% improvement over the source-only model, it relies on text data associated with the videos, which requires manual effort and contradicts our assumption. In contrast, our proposed method, \method{}, achieves a 9.76\% improvement over the source-only model without needing any text data in the target domain.
\begin{figure*}[t]
    \centering
    \includegraphics[width=0.9\textwidth]{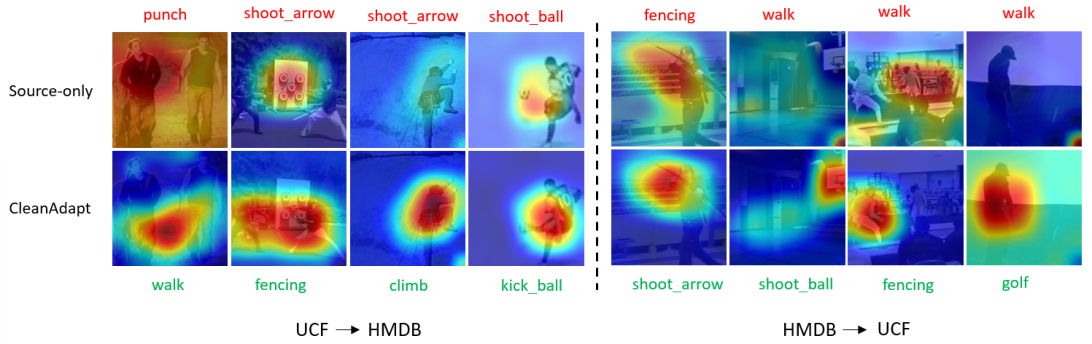}
\caption{Class activation map (CAM) on target-domain videos of the UCF $\leftrightarrow$ HMDB dataset. The actions in \textcolor{green}{green} are correct predictions, while actions in \textcolor{red}{red} are incorrect. It is worth noting that the adapted model in the bottom row emphasizes the action component rather than the contextual scene aspect. (Best viewed in color.)}
    \label{fig:heatmap}
\end{figure*}

\textbf{Visualization.} In Figure~\ref{fig:heatmap}, we show the Class Activation MAP (CAM) visualizations of our adapted model and compare them with the source-only model. The visualization shows that the source-only model attends to the background of the scene and makes incorrect predictions, while the adapted model focuses on the action component of the video to make correct predictions.

\setlength{\tabcolsep}{4pt}
\begin{table*}[h]
% \scriptsize
\centering
\caption{Performance comparison with zero-shot vision-language models on EPIC-Kitchens dataset. Results with zero-shot vision-language models are highlighted in \colorbox{lightcyan}{cyan} whereas the results our proposed model are highlighted in \colorbox{lightgray}{gray}.}
\resizebox{\textwidth}{!}{
\begin{tabular}{llllllll}
\hline
 \textbf{Method} & \textbf{D2$\rightarrow$D1} & \textbf{D3$\rightarrow$D1} & \textbf{D1$\rightarrow$D2} & \textbf{D3$\rightarrow$D2} & \textbf{D1$\rightarrow$D3} & \textbf{D2$\rightarrow$D3} & \textbf{Mean}   \\ \hline 

\rowcolor{LightCyan}

\textbf{Zero-shot Baselines} & &  &   & &  &  &   \\ \rowcolor{LightCyan}
 Video-LLaVA~\cite{lin2023video} &32.64&32.64 & 33.73 &33.73 & 34.90  & 34.90  &  33.76  \\ \rowcolor{LightCyan}
 EgoVLP~\cite{lin2022egocentric} &23.60 & 23.60 & 28.00  & 28.00 & 30.08 & 30.08 &  27.22  \\ \rowcolor{LightCyan}
 %EgoVLPv2~\cite{pramanick2023egovlpv2} & &  &   & &  & &    \\ \rowcolor{LightCyan}
LaViLA~\cite{zhao2023learning} &46.44 & 46.44 &  50.13 & 50.13 &47.64 &47.64  & 48.07  \\ \rowcolor{LightCyan}
 \rowcolor{Gray}
 \textbf{Source-free Domain Adaptation} & &  &   & &  &  &   \\ \rowcolor{Gray}
 Source only &41.8 & 41.1 & 41.9  & 46.1 & 37.3 & 43.9 &  42.0  \\ \rowcolor{Gray}
 \methodj{} & 48.3\textcolor{ForestGreen}{$\blacktriangle$ +6.5}& 48.7 \textcolor{ForestGreen}{$\blacktriangle$ +7.6} & 49.9 \textcolor{ForestGreen}{$\blacktriangle$ +8.0} & 56.3 \textcolor{ForestGreen}{$\blacktriangle$ +10.2}& 44.6 \textcolor{ForestGreen}{$\blacktriangle$ +7.3} &  48.9 \textcolor{ForestGreen}{$\blacktriangle$ +5.0}&  49.6 \textcolor{ForestGreen}{$\blacktriangle$ +7.6}  \\ \rowcolor{Gray}
 Target Supervised &62.3 & 62.3 & 72.7  & 72.7  & 71.1 & 71.1  & 68.4 \\ \hline
\end{tabular}}
\label{tab:zero-shot}
\end{table*}

\textbf{Comparisons with Zero-Shot Vision-Language Models.}
In addition to the state-of-the-art video domain adaptation approaches, we also compare our methods \method{} and \methodj{} with the following pre-trained vision-language models.

\textbf{Video-LLaVA~\cite{lin2023video}.} These baselines integrate visual representations into the language feature space, contributing to the development of unified Large Vision-Language Models (LVLMs). We prompt this LVLM with the following text - ``\textit{USER: $<video>$ Pick the action being performed in the video from the list below: [`take', `put', `open', `close', `wash', `cut', `mix', `pour'] ASSISTANT:}".

\textbf{EgoVLP~\cite{lin2022egocentric}.} 
A zero-shot baseline leverages the video-language pre-trained backbone from the Ego4D~\cite{grauman2022ego4d} dataset. We compute the embeddings for the \textit{class names} using the \textit{distilbert-base-uncased} model, and similarly, we extract video embeddings using the Ego4D~\cite{grauman2022ego4d} pre-trained backbone. Finally, the class name with the highest similarity is selected.

\textbf{LaViLA~\cite{zhao2023learning}.} This zero-shot baseline leverages the video-language pre-training by making a large-language model (LLM) conditioned on the visual inputs using the cross-attention layer on the Ego4D~\cite{grauman2022ego4d} dataset. Following~\cite{zhao2023learning}, we compute the embeddings for the \textit{class names} using the \textit{distilbert-base-uncased} model, and similarly, we extract video embeddings using the Ego4D~\cite{grauman2022ego4d} pre-trained backbone. Finally, the class name with the highest similarity is selected.

The results of the zero-shot baselines on the EPIC-Kitchens dataset are presented in Table~\ref{tab:zero-shot}. From the table, it is evident that zero-shot vision-language models do not perform particularly well on egocentric video recognition tasks. For instance, Video-LLaVA~\cite{lin2023video}, which is trained on generic third-person videos, achieves only 33.76\% average top-1 accuracy. In contrast, EgoVLP~\cite{lin2022egocentric}, and LaViLA~\cite{zhao2023learning}, pre-trained on egocentric videos from the Ego4D~\cite{grauman2022ego4d} dataset, outperform Video-LLaVA but still fall short by 22.4\% and 1.53\% when compared with our proposed method \methodj{}.

% \begin{figure*}[h]
% \begin{subfigure}{.49\textwidth}
%   \centering
%   % include first image
%   \includegraphics[width=0.7\linewidth]{assets/uh-ablation.png} 
%   \caption{UCF $\leftrightarrow$ HMDB}
%   \label{fig:ucf-hmdb}
% \end{subfigure}
% \begin{subfigure}{.49\textwidth}
%   \centering
%   % include second image
%   \includegraphics[width=0.7\linewidth]{assets/epic-ablation.png}  \caption{EPIC-Kitchens}
%   \label{fig:epic}
% \end{subfigure}
% \caption{Hyperparameter search for the value of keep-rate $\tau$ for UCF101$\leftrightarrow$ HMDB51 and EPIC-Kitchens dataset. The keep-rate $\tau$ controls the number of samples to be selected as clean having low-loss values computed against the pseudo-labels generated by the source-only model. All results reported here are for two-stream network. (Best viewed in color.)}
% \label{fig:split-ratio}
% \end{figure*}

% \begin{figure}[ht!]
%     \centering
%     \includegraphics[width=0.9\columnwidth]{assets/uh-ablation.png}
%     \includegraphics[width=0.9\columnwidth]{assets/epic-ablation.png}
%     \caption{Hyperparameter search for the value of keep-rate $\tau$ for UCF101$\leftrightarrow$ HMDB51 and EPIC-Kitchens datasets. The keep-rate $\tau$ controls the number of samples to be selected as clean having low-loss values computed against the pseudo-labels generated by the source-only model. All results reported here are for two-stream network. (Best viewed in color.)}
%     \vspace{-0.5cm}
%     \label{fig:split-ratio}
% \end{figure}

\begin{figure}[h]
\begin{subfigure}{.49\textwidth}
  \centering
  % include first image
  \includegraphics[width=0.9\linewidth]{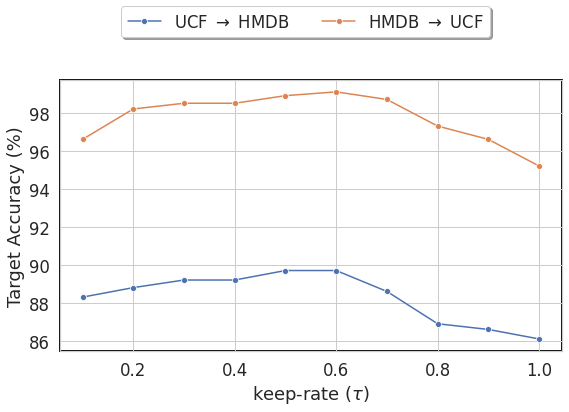} 
  \caption{UCF $\leftrightarrow$ HMDB}
  \label{fig:ucf-hmdb}
\end{subfigure}
\begin{subfigure}{.49\textwidth}
  \centering
  % include second image
  \includegraphics[width=0.9\linewidth]{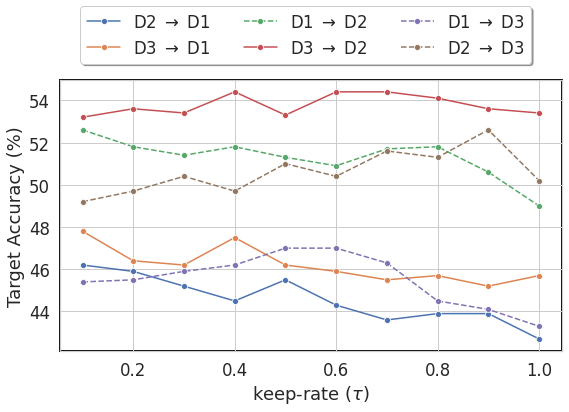}  \caption{EPIC-Kitchens}
  \label{fig:epic}
\end{subfigure}
\caption{Hyperparameter search for the value of keep-rate $\tau$ for UCF101$\leftrightarrow$ HMDB51 and EPIC-Kitchens dataset. The keep-rate $\tau$ controls the number of samples to be selected as clean due to low-loss values computed against the pseudo-labels. The results reported here are for the two-stream network. (Best viewed in color.)}
\label{fig:split-ratio}
\end{figure}

\subsection{Hyperparameter Search}\label{sssec:hs} The only hyperparameter our model introduces is the keep-rate $\tau$. It controls the number of target domain samples to be chosen from each class with low loss values in the adaptation stage. Figure~\ref{fig:split-ratio} shows the ablation results of varying $\tau$ in terms of validation accuracy for the target domain.

Empirically, we verify that the choice of keep-rate $\tau$ is important. As mentioned earlier, the samples from the target domain {\fontfamily{qcr}\selectfont train} set pseudo-labeled by the source-only model have inherently noisy labels. The choice of keep-rate $\tau = 1$ is equivalent to choosing all the samples for retraining the model on the target domain. However, the noisy pseudo-labels lead to a sub-optimal adaptation performance for all the datasets. For example, the adapted model gives top-1 accuracy of 86.1\% on UCF $\rightarrow$ HMDB and 95.2\% on HMDB $\rightarrow$ UCF respectively when $\tau$ is set to 1. However, when keep-rate $\tau$ is set to 0.6 gives top-1 accuracy of 89.8\% and 99.2\% on UCF $\rightarrow$ HMDB and HMDB $\rightarrow$ UCF respectively. 
\begin{figure}[h]
% \begin{subfigure}{\columnwidth}
  \centering
%   % include first image
  \includegraphics[width=0.6\textwidth]{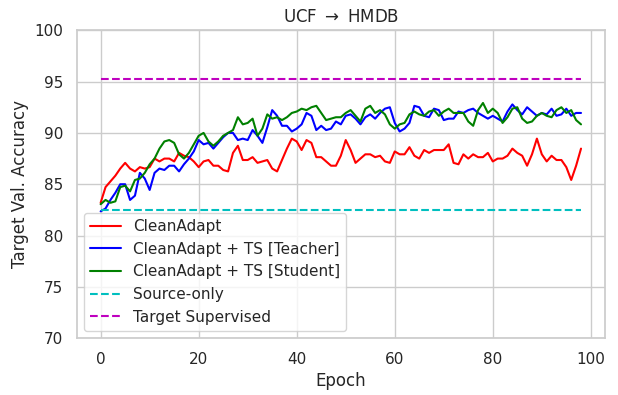} 
%   \caption{UCF $\rightarrow$ HMDB}
%   \label{fig:ucf-hmdb}
% \end{subfigure}
% \begin{subfigure}{\columnwidth}
%   \centering
%   % include second image
%   \includegraphics[width=0.9\columnwidth]{assets/hmdb_ucf_ts.png}  \caption{HMDB $\rightarrow$ UCF}
%   \label{fig:ts}
% \end{subfigure}

\caption{Analysis of the effect of \methodj{} on the target validation accuracy as compared to source-only, \method{}, and target-supervised methods. (Best viewed in color.)}
\label{fig:ts-exp}
\end{figure}
\subsection{Impact of Teacher-Student Framework}
\label{sec:ts}
To assess the effectiveness of the teacher-student framework, we generate a plot of target validation accuracy for each epoch during the training phase. We compare the performance of the source-only model, \method{}, and \methodj{}, and target-supervised models. Figure~\ref{fig:ts-exp} clearly demonstrates that the teacher-student framework contributes to stable pseudo-label generation, resulting in improved adaptation performance. To be precise, the teacher model serves as a regularizing factor for the student model, accomplishing this by producing consistent pseudo-labels. Consequently, the student model is guided to make gradual changes rather than abrupt ones. These pseudo-labels are treated by the student model as actual labels and take strongly augmented videos to generate robust features.
\begin{figure*}[h]
    \centering
    \includegraphics[width=0.9\textwidth]{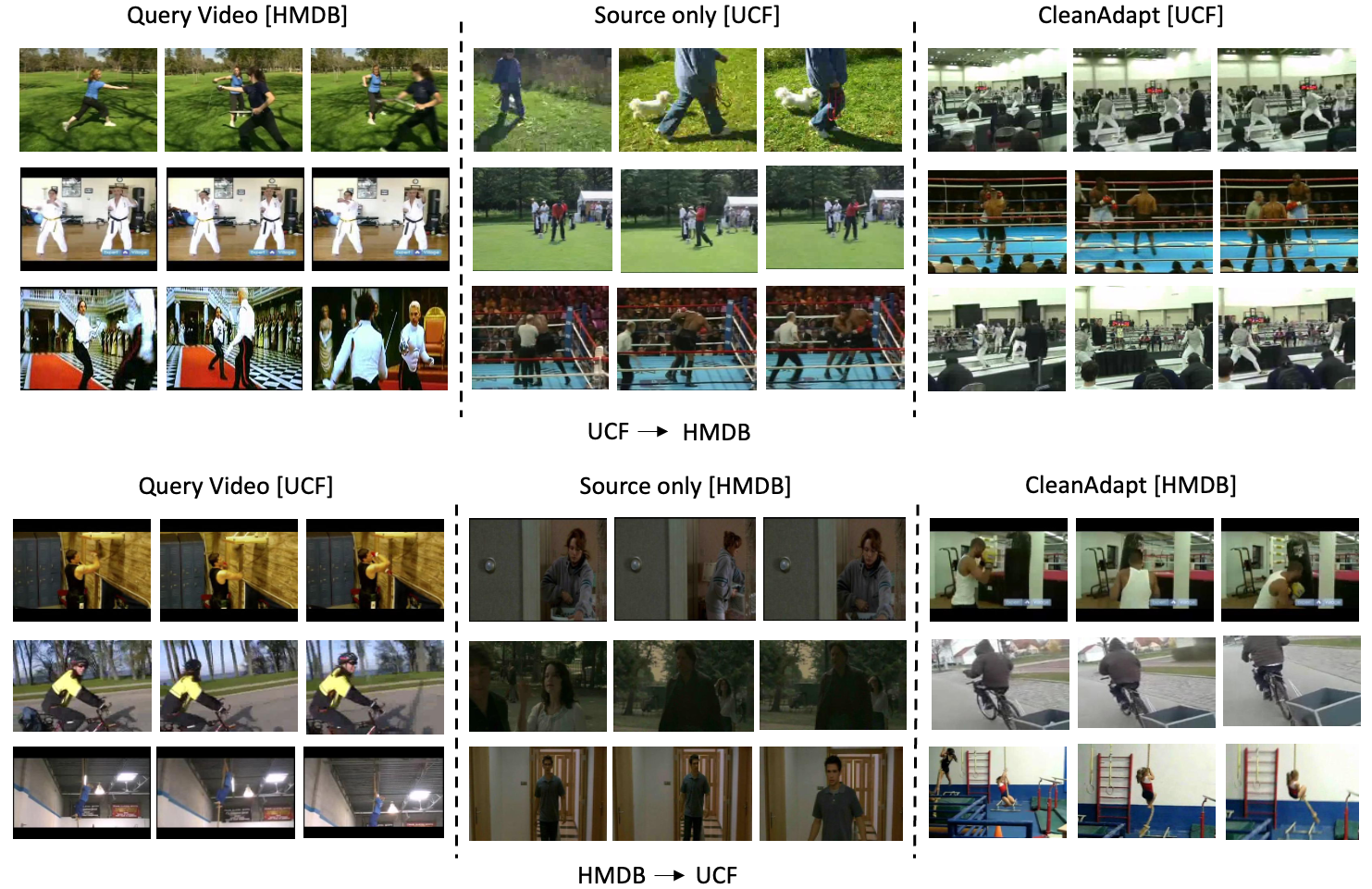}
    \caption{Nearest neighbour retrieval results for the UCF $\rightarrow$ HMDB and the HMDB $\rightarrow$ UCF dataset. The left column shows the query videos from the target domain. The middle column shows the retrieved source videos using the source-only model, and the right column shows the source videos retrieved using our proposed model. (Best viewed in color.)}
    \label{fig:retrieval}
\end{figure*}

% \begin{table}[t]
% \centering
% \scriptsize
% \caption{Cross-domain video retrieval results on UCF $\leftrightarrow$ HMDB dataset. Given queries from the target domain, we evaluate retrieved videos from the source domain in terms of R@k, where $k \in \{1, 5, 10\}$. Note that, all models reported here are two-stream networks and we average the similarity score from each modality to retrieve the source videos.}
% \begin{tabular}{ccccccc}
% \hline
% \multicolumn{1}{c}{\multirow{2}{*}{\textbf{Method}}} & \multicolumn{3}{c}{\textbf{UCF $\rightarrow$ HMDB }}                                              & \multicolumn{3}{c}{\textbf{HMDB $\rightarrow$ UCF}}                    \\ \cline{2-7} 
% \multicolumn{1}{c}{}                         & \multicolumn{1}{c}{R@1} & \multicolumn{1}{c}{R@5} & \multicolumn{1}{c}{R@10} & \multicolumn{1}{c}{R@1} & R@5 & \multicolumn{1}{c}{R@10} \\ \hline
%             Source Only                                 &     0.82                    &        0.87                 &         0.90                  &       0.88                  &   0.94  &     0.95                     \\\rowcolor{Gray}
%               \method{}                               &      \textbf{0.92}                   &          \textbf{0.97 }              &          \textbf{ 0.99 }               &        \textbf{0.91}                 &  \textbf{ 0.97}  & \textbf{0.98 }                         \\ \hline
% \end{tabular}
%  \label{tab:uh-retrieval}
% \end{table}

\begin{table}[ht]
\centering
\scriptsize
\caption{Cross-domain video retrieval results on UCF $\leftrightarrow$ HMDB dataset. Given queries from the target domain, we evaluate retrieved videos from the source domain in terms of R@k, where $k \in \{1, 5, 10\}$. Note that, all models reported here are two-stream networks and we average the similarity score from each modality to retrieve the source videos.}
\begin{tabular}{@{}lrrrrrr@{}}
\toprule

{\multirow{2}{*}{\textbf{Method}}} & \multicolumn{3}{c}{\textbf{UCF $\rightarrow$ HMDB}} & \multicolumn{3}{c}{\textbf{HMDB $\rightarrow$ UCF}}\\

\cmidrule(lr){2-4}\cmidrule(l){5-7}

\multicolumn{1}{c}{} & \multicolumn{1}{c}{R@1} & \multicolumn{1}{c}{R@5} & \multicolumn{1}{c}{R@10} & \multicolumn{1}{c}{R@1} & R@5 & \multicolumn{1}{c}{R@10} \\

\midrule

Source Only & $0.82$ & $0.87$ & $0.90$ & $0.88$ & $0.94$  & $0.95$ \\

\method{} & $\textbf{0.92}$ & $\textbf{0.97}$ & $\textbf{0.99}$ & $\textbf{0.91}$ & $\textbf{0.97}$  & $\textbf{0.98}$ \\

\bottomrule

\end{tabular}
 \label{tab:uh-retrieval}
\end{table}

\setlength{\textfloatsep}{0.3cm}

\subsection{Cross-domain Video Retrievals}
\label{sec:retrieval}
We examine the feature space learned by our adapted model \method{} to gain insights into its predictions through cross-domain video retrieval performance. Given a query video of a particular class from the target domain, we aim to retrieve videos from the source domain with the same semantic category. We show the results for the two-stream networks, where we first compute the similarity scores for the individual modalities and average them for final retrieval. We evaluate both the source-only and the proposed method \method{} quantitatively as well as qualitatively.

In Table~\ref{tab:uh-retrieval}, we show the quantitative results for the cross-domain video retrieval task for the UCF $\leftrightarrow$ HMDB dataset. Our model retrieves better source videos from the target queries with R@1 of 0.92 and 0.91 as compared to the source-only model, which achieves only 0.82 and 0.88 on UCF $\rightarrow$ HMDB and HMDB $\rightarrow$ UCF datasets respectively. In Figure~\ref{fig:retrieval}, we show qualitative retrieval results for the UCF $\rightarrow$ HMDB. Our model can correctly retrieve the source videos of the same semantic categories as the target query videos. 

\begin{table*}[h!]

\centering
\caption{Cross-domain video retrieval results on the EPIC-Kitchens dataset. Given queries from the target domain, we evaluate retrieved videos from the source domain in terms of R@k, where $k \in \{1, 5, 10\}$. All the models reported here are two-stream networks and we average the similarity score from each modality to retrieve the source videos.}
\resizebox{\textwidth}{!}{
\begin{tabular}{@{}lrrrrrrrrrrrrrrrrrr@{}}
\toprule
\multicolumn{1}{c}{\multirow{2}{*}{\textbf{Method}}} & \multicolumn{3}{c}{\textbf{D2 $\rightarrow$ D1 }} & \multicolumn{3}{c}{\textbf{D3 $\rightarrow$ D1}}   & \multicolumn{3}{c}{\textbf{D1 $\rightarrow$ D2}}   &  \multicolumn{3}{c}{\textbf{D3 $\rightarrow$ D2}}&

\multicolumn{3}{c}{\textbf{D1 $\rightarrow$ D3}}& \multicolumn{3}{c}{\textbf{D2 $\rightarrow$ D3}}\\

\cmidrule(lr){2-4}\cmidrule(lr){5-7}\cmidrule(lr){8-10}\cmidrule(lr){11-13}\cmidrule(lr){14-16}\cmidrule(l){17-19}

\multicolumn{1}{c}{} & \multicolumn{1}{c}{R@1} & \multicolumn{1}{c}{R@5} & \multicolumn{1}{c}{R@10} & \multicolumn{1}{c}{R@1} & R@5 & \multicolumn{1}{c}{R@10}& \multicolumn{1}{c}{R@1} & \multicolumn{1}{c}{R@5} & \multicolumn{1}{c}{R@10} & \multicolumn{1}{c}{R@1} & \multicolumn{1}{c}{R@5} & \multicolumn{1}{c}{R@10} &\multicolumn{1}{c}{R@1} & \multicolumn{1}{c}{R@5} & \multicolumn{1}{c}{R@10} &\multicolumn{1}{c}{R@1} & \multicolumn{1}{c}{R@5} & \multicolumn{1}{c}{R@10}\\\midrule

Source only & $0.35$ & $0.65$&$0.77$ & \textbf{$0.38$} & $0.68$&$0.79$ & $0.35$ & $\textbf{0.75}$& $0.86$& $0.41$ & $\textbf{0.77}$ & $0.84$ &$0.34$ &$0.68$&$0.82$ & $\textbf{0.42}$& $\textbf{0.74}$ & $\textbf{0.84}$ \\
\method{} & $\textbf{0.42}$& $\textbf{0.68}$ & $\textbf{0.80}$ & $0.37$&$\textbf{0.75}$ &$\textbf{0.83}$ & $\textbf{0.42}$&$0.74$ &$\textbf{0.87}$ & $\textbf{0.46}$ &$\textbf{0.77}$ &$\textbf{0.85}$ & $\textbf{0.35}$ &$\textbf{0.69}$ &$\textbf{0.83}$ & $0.40$ & $0.70$ & $0.82$ \\ \bottomrule
          
\end{tabular}}
 \label{tab:epic-retrieval}
 \vspace{-0.2cm}
\end{table*}
As shown in Table~\ref{tab:epic-retrieval}, our proposed approach achieves better retrieval performance in most of the cases than the source-only model for the EPIC-Kitchen dataset. Only for D2 $\rightarrow$ D3, our model under-performs the source-only model. This can be attributed to the fact that our model does not use source data during the adaptation stage, and thus, the model might start forgetting some attributes of the source dataset. 

\subsection{Limitations and Future Work}
% \textcolor{red}{In addition to the experiments described above, we conducted further experiments on more challenging datasets, including CharadesEgo~\cite{sigurdsson2018charades} and Ego2Exo~\cite{kalluri2024lagtran} to study the failure modes of our proposed approach.
% }

% \textbf{CharadesEgo.}

In our model, we do not explicitly incorporate the spatiotemporal relationships inherent in videos. This information is crucial for capturing how objects and actions evolve over time, which can significantly enhance the model's ability to understand complex video content. Without modeling these relationships, the model may struggle to effectively adapt to more challenging tasks such as video object segmentation, where it is essential to accurately track and delineate objects across frames.

A key assumption of our method is that each sample in both the source and target domains is associated with a single label and that they share a common label space $C$. While this is a standard setup in the literature, it may not always apply. For instance, CharadesEgo~\cite{sigurdsson2018charades} contains both first- and third-person videos, with each video having multiple labels. Our proposed approaches cannot be directly applied to such scenarios, and extending them to handle multi-label settings is left as future work.

However, we believe that the fundamental concept of selecting clean samples from noisy pseudo-labeled data will still be advantageous for tasks beyond video domain adaptation.
%-----------Conclusion-----------

\section{Conclusion}
\label{sec:conclusion}

In this work, we address the relatively under-explored problem of source-free video domain adaptation and propose two simple yet effective approaches: \method{} and \methodj{}. Our framework is based on self-training in which we generate noisy pseudo-labels for the target domain data using the source pre-trained model. Moreover, if we can filter out the noisy samples with varying $\tau$ using our proposed approach and use only the clean samples for fine-tuning, we achieve state-of-the-art performance without any video-specific modeling. To mitigate this issue of noisy pseudo-labels impeding the adaptation performance, we leverage the deep memorization effect~\cite{arpit2017closer} to identify and select the clean samples. 
Furthermore, we demonstrate that the quality of the clean samples can be improved by introducing a teacher-student framework, which in turn enhances the overall reliability of the adaptation training process and results in further performance improvements. Our methods consistently outperform recent image-based and video-based UDA methods without any source domain videos, thus establishing a new state of the art across several benchmarks.

While it is true that we get approximately $90\%$ accurate pseudo-labels for the HMDB $\xrightarrow{}$ UCF task, we want to emphasize that this is not always the case. For instance, in the EPIC-Kitchens dataset, the source-only model generates pseudo-labels with only about 42.0\% \textit{clean samples}. As demonstrated in Table~\ref{tab:epic} and Figure~\ref{fig:split-ratio}, training with all pseudo-labels without filtering the clean samples leads to sub-optimal performance. This underscores the importance of selecting an appropriate keep-rate ($\tau$). We would like to emphasize that the success of our method relies on the noise level in the pseudo-labeled target domain samples and the network's ability to avoid memorizing them. These might fail when the amount of noise is too much (e.g., 90\% noisy samples). Nonetheless, as demonstrated empirically, our method is effective even with a reasonable noise level.

\vspace{-0.5cm}
\section*{Acknowledgements} \noindent Avijit Dasgupta is supported by a Google Ph.D.\ India Fellowship. Karteek Alahari is supported in part by the ANR grant AVENUE (ANR-18-CE23-0011).

\bibliographystyle{elsarticle-num-names}
\bibliography{reference}
\end{document}